\documentclass[sigconf,anonymous=False]{acmart}

\AtBeginDocument{%
  \providecommand\BibTeX{{%
    \normalfont B\kern-0.5em{\scshape i\kern-0.25em b}\kern-0.8em\TeX}}}
    
\usepackage{graphicx}
\usepackage{subcaption}
\usepackage{wrapfig}
\usepackage{dsfont}
\usepackage{textgreek}
\usepackage{scalerel}
\usepackage{subcaption}
\usepackage{algorithm}
\usepackage{algpseudocode}
\usepackage{scalerel}
\usepackage{amsmath}
\usepackage{calc}
\usepackage{enumitem}

\setcopyright{acmcopyright}

\acmConference[Preprint]{}{currently under review.}{}

\setcopyright{licensedothergov}\acmConference[Preprint]{}{version}{currently under review.}

\begin{document}

\title{Federated and Privacy-Preserving Learning of \\ Accounting Data in Financial Statement Audits}

\author{Marco Schreyer}
\affiliation{%
    \institution{University of St\@.Gallen (HSG)}
    \city{St. Gallen}
    \country{Switzerland}
}
\email{marco.schreyer@unisg.ch}

\author{Timur Sattarov}
\affiliation{%
    \institution{Deutsche Bundesbank}
    \city{Frankfurt am Main}
    \country{Germany}
}
\email{timur.sattarov@bundesbank.de}

\author{Damian Borth}
\affiliation{%
    \institution{University of St\@.Gallen (HSG)}
    \city{St. Gallen}
    \country{Switzerland}
}
\email{damian.borth@unisg.ch}


\newcommand{\Tau}{\mathrm{T}}

\renewcommand{\shortauthors}{Schreyer, et al.}

\begin{abstract}


The ongoing `digital transformation' fundamentally changes audit evidence's nature, recording, and volume. Nowadays, the \textit{International Standards on Auditing} (ISA) requires auditors to examine vast volumes of a financial statement's underlying digital accounting records. As a result, audit firms also `digitize' their analytical capabilities and invest in \textit{Deep Learning} (DL), a successful sub-discipline of \textit{Machine Learning}. The application of DL offers the ability to learn specialized audit models from data of multiple clients, e.g., organizations operating in the same industry or jurisdiction. In general, regulations require auditors to adhere to strict data confidentiality measures. At the same time, recent intriguing discoveries showed that large-scale DL models are vulnerable to leaking sensitive training data information. Today, it often remains unclear how audit firms can apply DL models while complying with data protection regulations. In this work, we propose a \textit{Federated Learning} framework to train DL models on auditing relevant accounting data of multiple clients. The framework encompasses \textit{Differential Privacy} and \textit{Split Learning} capabilities to mitigate data confidentiality risks at model inference. Our results provide empirical evidence that auditors can benefit from DL models that accumulate knowledge from multiple sources of proprietary client data.

\end{abstract}



\maketitle

\section{Introduction}
\label{sec:introduction}

\begin{figure}[t!]
    \vspace{3mm}
	\hspace*{0.0cm} \includegraphics[width=8.2cm, angle=0, trim={1.0cm 0.0cm 0.0 0.0}]{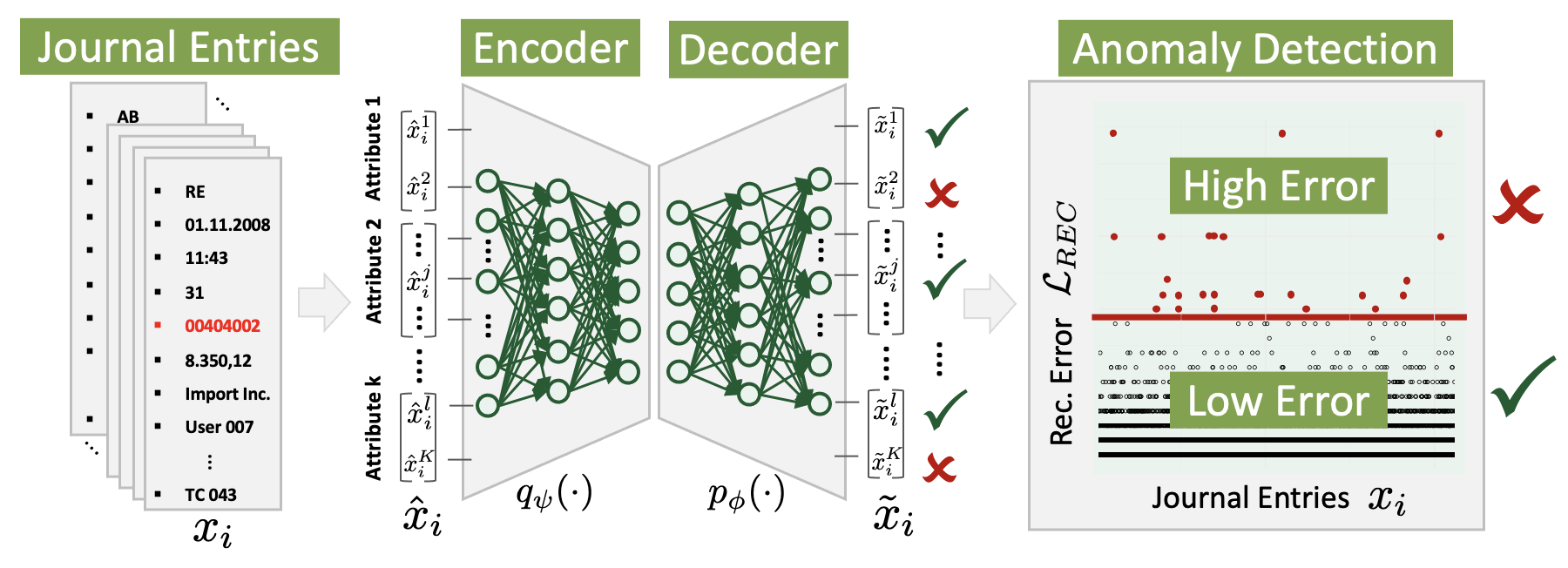}
	\vspace{-3mm}
	\caption{Schematic view of a deep \textit{Autoencoder Network} (AEN) based approach to detect accounting anomalies in large-scale journal entry data as proposed in \cite{schreyer2017}.}
	\label{fig:vis_anomaly_detection_setup}
	\vspace{-3mm}
\end{figure}

Nowadays, large audit firms often audit multiple clients in the same industry \cite{hoitash2006}. In general, auditors are not restricted to using proprietary client data to improve the quality of their assurance services \cite{kogan2021}.\footnote{Interpretations of the Statements of Auditing Standard (SAS) No. 56 indicate that `in circumstances where the auditor specializes in a specific industry, the auditor may use client data' to develop reasonable expectations \cite{guy2002}.} As a result, audit firms invest in the accumulation of industry-specific expertise to enhance their analytical capabilities and create synergies \cite{hogan1999}. This initiative originates from the assumption that audit clients operating in the same industry are affected by similar economic and societal factors, e.g., competitors, supply chains, fiscal policy, or legislation \cite{chan2004}. In parallel, the ongoing `digital transformation' fundamentally changed the nature, recording, and volume of audit evidence \cite{yoon2015}. Nowadays, auditors audit vast quantities of accounting transactions \cite{ISA240}, referred to as \textit{Journal Entries} (JEs), recorded in their client's \textit{Enterprise Resource Planning} (ERP) systems \cite{grabski2011}. The unprecedented availability of data creates a unique opportunity for the audit profession to derive data-driven insights \cite{appelbaum2016}. As a result, audit firms `digitize' their analytical capabilities to provide adequate assurance services \cite{alles2015}. In this context, audit firms also facilitate the application of \textit{Deep Learning} (DL) \cite{lecun2015}, a sub-discipline of \textit{Machine Learning} (ML) \cite{sun2019}, to learn complex models from digital audit evidence. Lately, such models have been trained to accomplish a various downstream audit tasks, e.g. accounting anomaly detection \cite{schultz2020, zupan2020, nonnenmacher2021a}, audit sampling \cite{schreyer2020, schreyer2021} or notes analysis \cite{sifa2019, ramamurthy2021}. Figure \ref{fig:vis_anomaly_detection_setup} illustrates the exemplary application of a deep bottleneck \textit{Autoencoder Network} (AEN) to detect anomalies in accounting data \cite{schreyer2017}. However, the learning of `specialist' or `meta' ML models across data from multiple audit clients is a promising but under-explored domain. 

\begin{figure*}[ht!]
    \begin{minipage}[b]{.40\linewidth}
    \center
    \includegraphics[height=5.8cm,trim=0 0 0 0, clip]{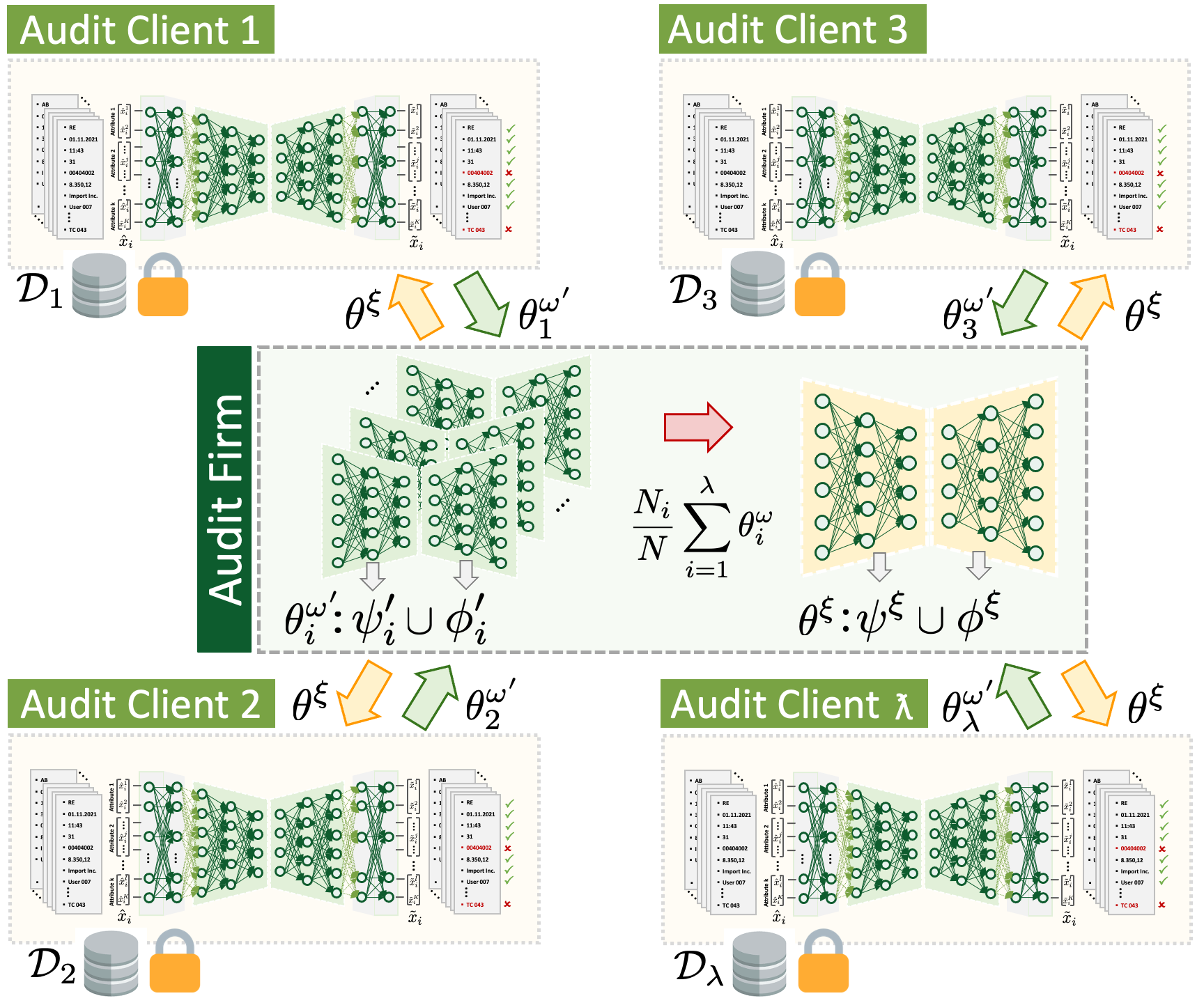}\\
    \vspace{1mm}
    \textbf{(a) Federated Training Setup}
    \end{minipage}
    \begin{minipage}[b]{.08\linewidth}
    \center
    \hspace{1mm}
    \end{minipage}
    \begin{minipage}[b]{.40\linewidth}
    \center
    \includegraphics[height=5.5cm,trim=0 0 0 0, clip]{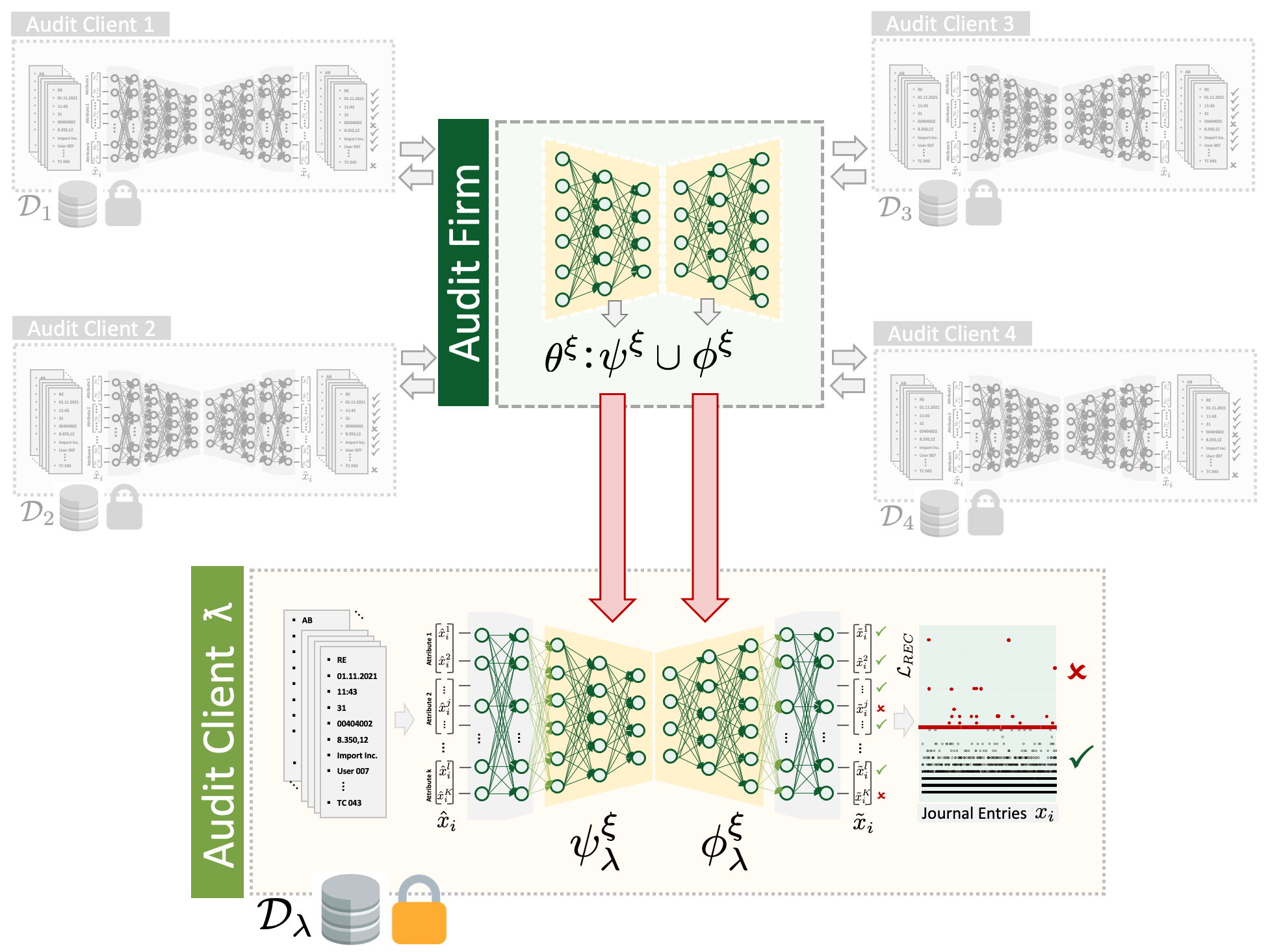}\\
    \vspace{1mm}
    \textbf{(b) Federated Inference Setup}
    \end{minipage}
    \vspace{-3mm}
    \caption{(a) The proposed federated learning approach to learn from confidential accounting data of multiple clients. At training time, the clients train \textit{decentral} AEN models with parameters \scalebox{0.8}{$\theta^{\omega}_{i}$} on their private datasets \scalebox{0.9}{$\mathcal{D}_{i}$}. Upon successful training, the audit firm aggregates the client models into a \textit{central} model with parameters \scalebox{0.8}{$\theta^{\xi}$}. (b) At inference time, the audit firm's central AEN model parameters \scalebox{0.8}{$\theta^{\xi}$} are broadcasted to a respective client to conduct audit procedures.}
    \label{fig:architecture}
    \vspace{-3mm}
\end{figure*}

In 2017 McMahan et al. \cite{mcmahan2017a} introduced the concept of \textit{Federated Learning} (FL) that enables distributed clients to collaboratively train models under the orchestration of a central server \cite{kairouz2019}. A key advantage of the FL setup is the idea of local model training and central model aggregation \cite{mathews2022}. Thereby, the training data remains decentralized, enabling the clients to learn a common model without sharing proprietary data \cite{konevcny2016}. However, as the benefits of such a learning scheme become apparent for audit firms, there is a gradual awakening to the fact that associated risks may also arise \cite{munoko2020}. Recently, it has been demonstrated that ML models are vulnerable to data leakage attacks, e.g., designed to extract sensitive or personally identifiable information \cite{carlini2020, yin2021}. 

At the same time, auditors are obliged to protect the confidentiality of their client's data \cite{AICPAEthics2014}.\footnote{Specifically, Rule 1.700.001.01 of the \textit{Americian Institute of Certified Public Accountants} (AICPA) Code of Professional Conduct \cite{AICPAEthics2014} states that \textit{`a member in public practice shall not disclose any confidential client information without the specific consent of the client'}. Following the AICPA's definition \textit{`confidential client information'} is defined as \textit{`any information obtained from the client that is not available to the public'}.} Currently, audit firms train ML models using proprietary client data, e.g., the journal entries posted in a client's ERP system by an employee exhibiting sensitive information. Furthermore, (inter-) national legislation requires auditors to implement measures that ensure the protection of data privacy.\footnote{For example, the U.S. \textit{Gramm Leach Bliley Act}, the \textit{California Consumer Privacy Act}, or the E.U. \textit{General Data Protection Regulation} enforce strict data privacy obligations.} Such regulations pose additional requirements for the cross-client learning and application of ML-based audit models. In 2006, Dwork et al. \citep{dwork2006a} proposed the concept of \textit{Differential Privacy} (DP) to address the challenge of inferring confidential training data information through ML models. One of the DP learning setup's central ideas is introducing random noise perturbations to ensure privacy through plausible deniability. Today, it often remains unclear how audit firms can use such learning setups and preserve the confidentiality of sensitive client data.

Inspired by these developments, we examine whether FL enhanced by DP can be utilized to (i) learn industry-specific audit models that (ii) comply with data confidentiality obligations. In summary, we present the following primary contributions:

\begin{itemize}

\item We propose a \textit{Federated Learning} framework that enables auditors to learn industry-specific models from data of multiple clients without the need to access it centrally.

\item We demonstrate how such a framework can be enhanced by \textit{Differential Privacy} and \textit{Split Learning} to learn confidentiality-preserving models of accounting data.

\item We conduct an extensive evaluation using real-world datasets and illustrate the benefits of the learning setup based on the general audit procedure of detecting anomalies.

\vspace{-0.0cm}

\end{itemize}

Driven by the progressive application of DL in financial audits, new and effective ways to securely learn `meta' ML models from client data become of high relevance \citep{yunis2021}. We see this work as a promising step toward the future adoption of \textit{Artificial Intelligence} (AI) augmented auditing by audit firms. The remainder of this work is structured as follows: In Section \ref{sec:relatedwork}, we provide an overview of related work. Section \ref{sec:methodology} follows with a description of the proposed framework to learn federated and privacy-preserving models from vast quantities of JE data. The experimental setup and results are outlined in Section \ref{sec:experimental_setup} and Section \ref{sec:experimental_results}. In Section \ref{sec:summary}, the paper concludes with a summary of the work. 

\vspace{-2mm}

\section{Related Work}
\label{sec:relatedwork}

The application of ML in financial audits triggered a sizable body of research by academia \citep{appelbaum2016, sun2019} and practitioners \citep{sun2017, dickey2019}. In this section, we present our literature study focusing on the federated and differential-private learning of accounting data representations.

\vspace{0.5mm}

\textbf{Federated Learning} enables multiple decentral entities to collaboratively train ML models under the orchestration of a central trusted entity without compromising data protection \citep{kairouz2019}. Model aggregation is one of the most common federated learning mechanisms, which trains a global model by summarizing the model parameters from all decentral parties. McMahan et al. \cite{mcmahan2017a} proposed \textit{FedAvg} a deep network federation learning framework based on an iterative model averaging the local model in each round of updates. Papernot et al. \cite{papernot2016} introduced \textit{PATE}, which aggregates knowledge transferred from a teacher model that is trained on separated data of student models. Yurochkin et al. \cite{yurochkin2019} developed a Bayesian framework for federated learning, establishing a global model by matching neurons of local models. The combination of federated learning and multitasking \cite{smith2017} allows multiple users to train models of different tasks locally, which is also a typical method of model aggregation. In \cite{kim2018}, federated learning and blockchain are combined to exchange and update the model data of each piece of equipment based on the blockchain. Recently FL setups have been deployed in a variety of application scenarios, e.g., recommendation systems \cite{hard2018} and finance \cite{kawa2019}.

\vspace{0.5mm}

\textbf{Privacy-Preserving Learning} emerged to protect data processed by ML algorithms and prevent information leakage to non-trusted third parties. Nowadays, prominent techniques encompass secure multi-party computation \cite{yao1982}, homomorphic encryption \cite{rivest1978}, and differential privacy \cite{dwork2008}. Differential Privacy addresses the issue that attackers might infer sensitive training data information through trained models \cite{dwork2006a}. Dwork et al. \cite{dwork2014} initially proposed the \textit{Gaussian Mechanism} (GM) to solve the challenge of real-valued database queries. The idea of GM is the addition of noise to each dimension of the query result. Several approaches have been proposed in the context of ML to inject noise perturbations into the learning process. Fukuchi et al. \cite{fukuchi2017}, and Kang et al. \cite{kang2020} applied such perturbations to the input training data. In contrast, Zhang et al. \cite{zhang2017} introduced gradient descent learning with model output perturbations. In the context of training deep learning models, a significant body of research investigates the idea of designing deliberate gradient perturbations \cite{abadi2016}, as well as perturbations of the learning objective \cite{kifer2012}. In addition, Papernot et al. \cite{papernot2016} proposed an approach of adding perturbation-based noisy labelling. Nowadays, DP learning mechanisms are applied in sensitive areas, such as healthcare \cite{dankar2012} and financial risk modelling \cite{zheng2020}.



\vspace{0.5mm}

\textbf{Representation Learning} denotes techniques that allow systems to discover relevant data features to solve a specific task \cite{bengio2013}. Nowadays, the majority of ML methods used in financial audits depend on `human' engineered data features \cite{cho2020}. Such techniques encompass, for example, Naive Bayes classification \cite{Bay2002}, network analysis \cite{McGlohon2009}, univariate and multivariate attribute analysis \cite{Argyrou2012}, cluster analysis \cite{thiprungsri2011}, transaction log mining \cite{Khan2009}, or business process mining \cite{Jans2011}. With the advent of DL, novel representation learning techniques have emerged in financial audits \cite{nonnenmacher2021b}. Such methods rely on accounting data representations that are learned `end-to-end' without human intervention \cite{sun2019}. Nowadays, applied representation learning techniques encompass, autoencoder neural networks \cite{schultz2020}, adversarial autoencoders \cite{schreyer2019b}, or variational autoencoders \cite{zupan2020}. Lately, self-supervised learning techniques have been proposed to learn rich representations for multiple downstream audit tasks \cite{schreyer2021}.

\vspace{0.5mm}


\noindent Concluding the literature survey and, to the best of our knowledge, this work presents the first step towards learning accounting data representations in a confidentiality-preserving manner.

\section{Methodology}
\label{sec:methodology}

In this section, we introduce a framework for learning accounting data from multiple audit clients for accounting anomaly detection. The framework comprises four interacting learning setups, namely (i) \textit{representation}, (ii) \textit{federated}, (iii) \textit{split}, and (iv) \textit{differential-private} learning, described in the following.

\vspace{0.5mm}

\textbf{Accounting Journal Entries:} Formally, let \scalebox{0.9}{$\mathcal{D} = \{x_{1}, x_{2}, ..., x_{N}\}$} define a population of \scalebox{0.9}{$i=1, 2, 3, ..., N$} \textit{journal entries}. Each individual entry, denoted by \scalebox{0.9}{$x_{i} = \{x_{i}^{1}, x_{i}^{2}, ..., x_{i}^{M}; x_{i}^{1}, x_{i}^{2}, ..., x_{i}^{K}\}$}, consists of \scalebox{0.9}{$j=1, 2, 3, ..., M$} categorical accounting attributes and \scalebox{0.9}{$l=1, 2, 3, ..., K$} numerical accounting attributes. The individual attributes encompass the journal entry's details, such as an entry's posting type, posting date, amount, or general ledger. When examining real-world journal entry data populations, two characteristics can be observed: First, the journal entry attributes exhibit a high variety of distinct attribute values, e.g., due to the high number of vendors or distinct posting amounts. Second, journal entries exhibit a plethora of distinct attribute value correlations, e.g., a document type usually posted in combination with a specific general ledger and vendor. 

\vspace{0.5mm}

\textbf{Accounting Anomalies:} Derived from this observation and inspired by Breunig et al. \cite{Breunig2000}, we distinguish two types of `anomalous' entries \cite{schreyer2017}, namely \textit{global} and \textit{local} accounting anomalies:

\begin{itemize}

\item \textit{Global Accounting Anomalies} correspond to journal entries that exhibit unusual or rare individual attribute values, e.g., rarely used ledgers, or unusual posting times. Such anomalies often correspond to unintentional mistakes, are comparably simple to detect, and possess a high \textit{error} risk.

\item \textit{Local Accounting Anomalies} correspond to journal entries that exhibit unusual attribute value correlations, e.g., rare co-occurrences of ledgers and posting types. Such anomalies might correspond to intentional deviations, are comparably difficult to detect, and possess a high \textit{fraud} risk.

\end{itemize}

\noindent Ultimately, auditors aim to detect both types of accounting anomalies in real-world audits using unsupervised learning \cite{schultz2020, zupan2020, nonnenmacher2021a}.

\vspace{1mm}








\textbf{Representation Learning:} To establish the anomaly detection setup we utilize bottleneck \textit{Autoencoder Networks (AEN's)} \cite{hinton2006}. Formally, AEN's are comprised of two nonlinear functions referred to as \textit{encoder} and \textit{decoder} networks that are jointly trained to reconstruct a given input. The encoder network \scalebox{0.9}{$q_{\psi}(\cdot)$} maps an input \scalebox{0.9}{$x_{i} \in \mathcal{R}^k$} to a code vector \scalebox{0.9}{$z_{i} \in \mathcal{R}^m$} referred to as \textit{latent representation}, where usually \scalebox{0.9}{$k > m$}. Subsequently, the representation is mapped back by the decoder network \scalebox{0.9}{$p_\phi(\cdot)$} to a reconstruction \scalebox{0.9}{$\tilde{x}_{i} \in \mathcal{R}^k$} of the original input. The AENs parameter are denoted as \scalebox{0.9}{$\theta =\{\psi \cup \, \phi\}$}. In an attempt to achieve \scalebox{0.9}{$x \approx \tilde{x}$} both AEN networks are trained in parallel to minimize the dissimilarity of a given journal entry \scalebox{0.9}{$x_{i}$} and its reconstruction \scalebox{0.9}{$\tilde{x}_{i} = p_\phi(q_\psi(x_{i}))$} as faithfully as possible. Consequently, the AEN learns optimal model parameters \scalebox{0.9}{$\theta^{*} =\{\psi^{*} \cup \, \phi^{*}\}$} by optimizing a training objective, formally defined as:

\begin{equation}
    \arg \min_{\phi, \psi} \|x_{i} - p_{\phi}(q_{\psi}(x_{i}))\|.
    \vspace{-1mm}
    \label{equ:reconstruction_loss}
\end{equation}

\noindent Upon successful AEN training, we examine the reconstruction loss \scalebox{0.9}{$\mathcal{L}_{Rec}$} of each journal entry \cite{hawkins2002}. Entries that correspond to a high reconstruction error will be selected for detailed audit procedures \cite{schultz2020, schreyer2022}. Figure \ref{fig:vis_anomaly_detection_setup} illustrates a schematic view of the AEN based accounting anomaly detection approach.

\vspace{0.5mm}

\textbf{Federated Learning:} To preserve the confidentiality of client data, we extend the AENs anomaly detection capability by \textit{Federated Learning (FL)} \cite{mcmahan2017a}. We assume that the entire training data is available as \scalebox{0.9}{$K$} sub-datasets \scalebox{0.9}{$\mathcal{D}=\{\mathcal{D}_{i}\}^{K}_{i=1}$}. To establish the FL setup, let \scalebox{0.9}{$\xi$} denote a \textit{central} AEN model with parameters \scalebox{0.9}{$\theta^{\xi} =\{\psi^{\xi} \cup \, \phi^{\xi}\}$} maintained by a trusted audit firm. Let furthermore \scalebox{0.9}{$\{\omega\}^{K}_{i=1}$} denote a set of \textit{decentral} AEN models hosted by the audit clients. Formally, FL then solves an optimization task of the form: 

\begin{equation}
     \min_{x \in \mathcal{D}} \mathcal{L}(x) = \frac{1}{\lambda} \sum\nolimits_{i=1}^{\lambda} \mathcal{L}_{i}(x) \,,
    \vspace{2mm}
    \label{equ:parameter_aggregation}
\end{equation}

\noindent where \scalebox{0.9}{$\mathcal{L}_{i}(x) = \mathbb{E}_{\hat{x} \sim \mathcal{D}_{i}}[\omega_{i}(x, \hat{x})]$}, denotes the reconstruction loss of the \textit{i}-th client, and \scalebox{0.9}{$\mathcal{D}_{i}$} is the JE data subset of the \textit{i}-th client. Similar to real-world audits, each subset \scalebox{0.9}{$\mathcal{D}_{i}$} is only privately accessible by a single audit client. For \scalebox{0.9}{$i \neq j$}, \scalebox{0.9}{$\mathcal{D}_{i}$} and \scalebox{0.9}{$\mathcal{D}_{j}$} may be non-iid. An overview of the proposed training setup is illustrated in Fig. \ref{fig:architecture} (a). At training time, a synchronous update scheme is established that proceeds in \textit{rounds of communication} \cite{mcmahan2017a}. At each round $r \in R$, a set of \scalebox{0.9}{$\lambda$} of available audit clients \scalebox{0.9}{$\omega_{r} \subseteq \{\omega\}^{K}_{i=1}$} and \scalebox{0.9}{$|\omega_{r}|=\lambda$} is randomly selected. To initialize the FL, the auditor broadcasts its current \textit{central model} \scalebox{0.8}{$\theta^{\xi}_{r}$} to the selected clients. Subsequently, the clients \scalebox{0.9}{$\omega_{r}$} conduct a round of decentral training iterations using their private sub-datasets \scalebox{0.9}{$\mathcal{D}_{i}$}. At each client, the training results in an updated \textit{decentral model} \scalebox{0.9}{$\omega_{i}$} with parameters \scalebox{0.9}{$\theta^{\omega}_{i} =\{\psi^{\omega}_{i} \cup \, \phi^{\omega}_{i}\}$}. Upon successful training, the clients send their decentral model parameters \scalebox{0.8}{$\theta^{\omega}_{i,r}$} to the audit firm. Finally, the firm updates the central model parameters \scalebox{0.8}{$\theta^{\xi}_{r+1}$} by aggregating the received parameters. This process repeats until the final communication round \scalebox{0.9}{$R$} is reached. To aggregate the updates, we adapt the \textit{Federated Averaging (FedAvg)} \cite{mcmahan2017a} algorithm as presented in Algo \ref{alg:federated_averaging_algo}. \textit{FedAvg} computes a weighted average over the decentral updates, as defined by:

\begin{equation}
    \theta^{\xi}_{r+1} \leftarrow \frac{N_{i}}{N} \sum\nolimits_{i=1}^{\lambda} \theta^{\omega}_{i,r} \,,
    \vspace{2mm}
    \label{equ:parameter_aggregation}
\end{equation}

\noindent where $N$ denotes the total number of journal entries, and $N_{i}$ is the number of entries privately accessible by a particular client. The actual central model is then utilized at inference time to audit a client's proprietary data. An overview of the proposed inference scheme is illustrated in Fig. \ref{fig:architecture} (b).

\vspace{0.5mm}

\textbf{Split Learning:} When investigating the representations learned by AENs \cite{schreyer2019a} it becomes apparent that representations learned by early (later) layers of the encoder (decoder) function encode detailed attribute characteristics, e.g., a client's journal entry attribute co-occurrence pattern. In contrast, the later encoder (early decoder) function layers learn representations that encode general or `meta' posting patterns, e.g., a client's prevalent financial accounting principles \cite{schreyer2021, schreyer2022}. Derived from this observation and to increase the level of privacy, we apply principles of \textit{Split Learning (SL)} \cite{gupta2018}. Given an AEN architecture, we interpret the non-linear encoder (and decoder) function as a sequential application of layers, formally defined by \scalebox{0.9}{$q_{\psi}(\cdot) \leftarrow \ell_{N}^{q}(\ell_{N-1}^{q}...(\ell_{1}^{q}(\ell_{0}^{q}(\cdot))))$} (and \scalebox{0.9}{$p_{\phi}(\cdot) \leftarrow \ell_{N}^{p}(\ell_{N-1}^{p}...(\ell_{1}^{p}(\ell_{0}^{p}(\cdot))))$}). As illustrated in Fig. \ref{fig:vis_split_learning_setup}, we divide the encoder (decoder) network into two sub-networks: a \textit{private} \scalebox{0.8}{$q_{\psi}'$} and a \textit{public} sub-network \scalebox{0.8}{$q_{\psi}^{*}$} (\scalebox{0.85}{$p_{\phi}'$} and \scalebox{0.8}{$p_{\phi}^{*}$}). Thereby, the encoder (decoder) \textit{cut-layer} \scalebox{0.8}{$\ell_{c}^{q}$} (\scalebox{0.8}{$\ell_{c}^{p}$}) defines the network separation. In each communication round, a client only transfers its public parameters \scalebox{0.9}{$\theta^{\omega,*} = \{\psi^{*};\phi^{*}\}$} to the auditor. The private parameters \scalebox{0.9}{$\theta^{\omega'} = \{\psi';\phi'\}$} are not shared and kept private. 

\vspace{-1.0mm}

\algnewcommand\algorithmicinput{\textbf{Audit Firm executes:}}
\algnewcommand\AUDITOR{\item[\algorithmicinput]}

\algnewcommand\algorithmicoutput{\textbf{Audit Client executes:}}
\algnewcommand\AUDITCLIENT{\item[\algorithmicoutput]}

\begin{algorithm}
\caption{\textit{Federated Learning (FL)} setup inspired by McMahan et al. \cite{mcmahan2017a}, $\lambda$ denotes the number of audit clients, $\rho$ the batch-size, $\Tau$ the number of decentral iterations, and $\eta$ the learning rate.}
\label{alg:federated_averaging_algo}
\begin{algorithmic}[1]
\small
\AUDITOR \textit{// central}
\State randomly initialize $\theta^{\xi}$
\For{each round $r = 1, 2, ... \; R$}
    \State $\theta^{\omega}_{r} \gets$ select random set of $ \; \lambda \; $ clients
    
    \For{each client $\omega_{i} \in  \lambda \; \textbf{in parallel}$}
        \State $\theta^{\omega}_{i,r+1} \gets \text{ClientUpdate(i, $\theta^{\xi}_{r}$, r)}$
    \EndFor
    \State $\theta^{\xi}_{r+1} \gets  \frac{N_{i}}{N} \sum\nolimits_{i=1}^{\lambda} \theta^{\omega}_{i,r+1}$
\EndFor
\end{algorithmic}
\begin{algorithmic}[1]
\small
\AUDITCLIENT \textit{// decentral}
\State $\text{ClientUpdate(i, $\theta^{\xi}_{r}$, r):}$ \hspace{3mm}
    \State \hspace*{4mm} $\theta^{\omega}_{i,\tau} \gets$ initialize client model with $\theta^{\xi}_{r}$
    \State \hspace*{4mm} $\mathcal{B} \gets$ split private data $\mathcal{D}_{i}$ into batches of size $\rho$ each
    \State \hspace*{4mm} \textbf{for} each iteration $\tau = 1, 2, ... \; \Tau$ \textbf{do}
        \State \hspace*{8mm} $b_{i} \gets$ sample random batch $b \sim \mathcal{B}$
        \State \hspace*{8mm} $\theta^{\omega}_{i,\tau+1} \gets \theta^{\omega}_{i,\tau} - \eta \nabla \mathcal{L}^{Rec}(\theta^{\omega}_{i,\tau}, b_{i})$
    \State \hspace*{4mm} \textbf{end for}
    \State \hspace*{4mm} return $\theta^{\omega}_{i,\tau}$ to Audit Firm
\end{algorithmic}
\end{algorithm}

\begin{figure}[t!]
	\hspace*{0.0cm} \includegraphics[width=7.0cm, angle=0, trim={1.0cm 0.0cm 1.0cm 0.0cm}]{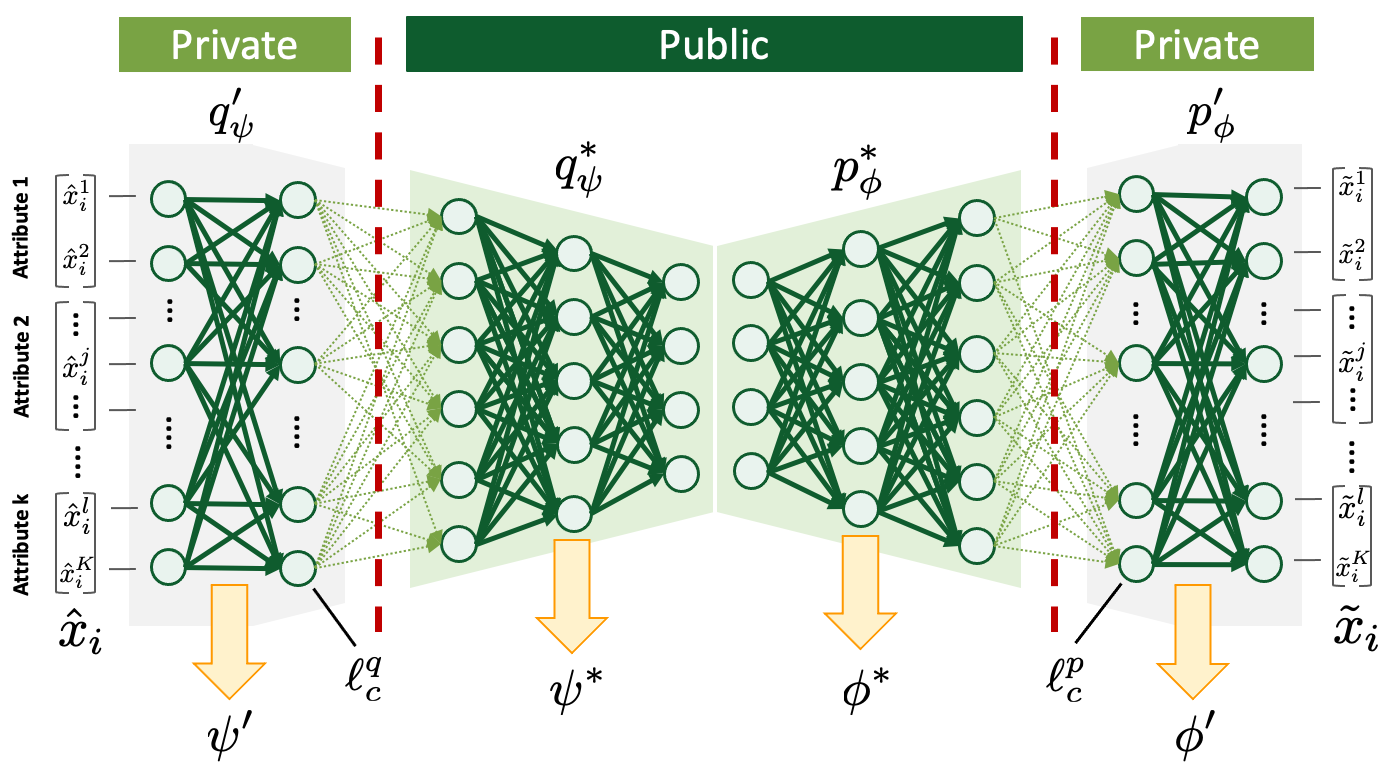}
	\vspace{-3mm}
	\caption{Proposed AEN \textit{Split Learning (SL)} \cite{gupta2018} setup dividing the decentral AEN models \scalebox{0.9}{$\{\omega\}^{K}_{i=1}$} into shared \textit{public} model parameters \scalebox{0.9}{$\{\psi_{i}';\phi_{i}'\}$} and \textit{private} model parameters \scalebox{0.9}{$\{\psi_{i}^{*};\phi_{i}^{*}\}$}.}
	\label{fig:vis_split_learning_setup}
	\vspace{-3mm}
\end{figure}

\vspace{1.0mm}


\textbf{Differential Private Learning:} To further reduce the risk of data leakage by the shared public model parameters \scalebox{0.9}{$\theta^{\omega, *}$} we apply \textit{Differential Privacy (DP)} \cite{dwork2006a} techniques. DP builds on the idea of introducing randomized noise into the learning procedure. A learning algorithm $A$ is understood to satisfy a $\epsilon-$DP mechanism if and only if for any pair of datasets $\mathcal{D}$ and $\mathcal{D}'$ that differ in only one element, the following statement holds true: 

\begin{equation}
    P[A(\mathcal{D}) = t] \leq e^{\epsilon} \, P[A(\mathcal{D}')=t], \forall t,
    \vspace{2mm}
    \label{equ:differential_privacy}
\end{equation}

\noindent where \scalebox{0.9}{$P[A(\mathcal{D}) = t]$} denotes the probability that algorithm \scalebox{0.9}{$A$} outputs \scalebox{0.9}{$t$} and \scalebox{0.9}{$\epsilon > 0$}. Thereby, the quantity \scalebox{0.9}{$ln \frac{P[A(\mathcal{D}) = t]}{P[A(\mathcal{D}') = t]}$} denotes the privacy loss. To establish \textit{Differential Private-Stochastic Gradient Descent (DP-SGD)} learning \cite{abadi2016}, we apply two modifications to the original SGD algorithm. First, the sensitivity of each per sample gradient is bounded by clipping it in the \textit{$l2$}-norm. We denote the parameter \scalebox{0.9}{$\nabla_{max}$} as the maximum norm threshold of the per-sample gradients. Second, before updating the model parameters, perturbation noise is added to the clipped gradients using the \textit{Gaussian Mechanism} (GM) \cite{dwork2014}. Assuming a target function \scalebox{0.9}{$f: \mathcal{D} \rightarrow \mathbb{R}$} over the data \scalebox{0.9}{$D$}, the GM formally denotes a randomization algorithm  \scalebox{0.9}{$A_{f}(\mathcal{D}) = f(\mathcal{D}) + z$}. Thereby,  \scalebox{0.9}{$z$} denotes a perturbation random variable drawn from a Gaussian distribution \scalebox{0.9}{$z \sim \mathcal{N}(\mu, \sigma)$}. The amount of added noise is determined by the Gaussian's standard deviation \scalebox{0.9}{$\sigma$} derived from \scalebox{0.9}{$\sigma = \nabla_{max} \times \kappa$}. To ensure $\epsilon-$DP, increasing the per sample gradient threshold \scalebox{0.9}{$\nabla_{max}$} will result in additional noise, where the parameter \scalebox{0.9}{$\kappa$} denotes a factor of noise multiplication.





\begin{figure*}[ht!]
    \begin{minipage}[b]{.32\linewidth}
        \center
        \includegraphics[height=4.6cm,trim=0 0 0 0, clip]{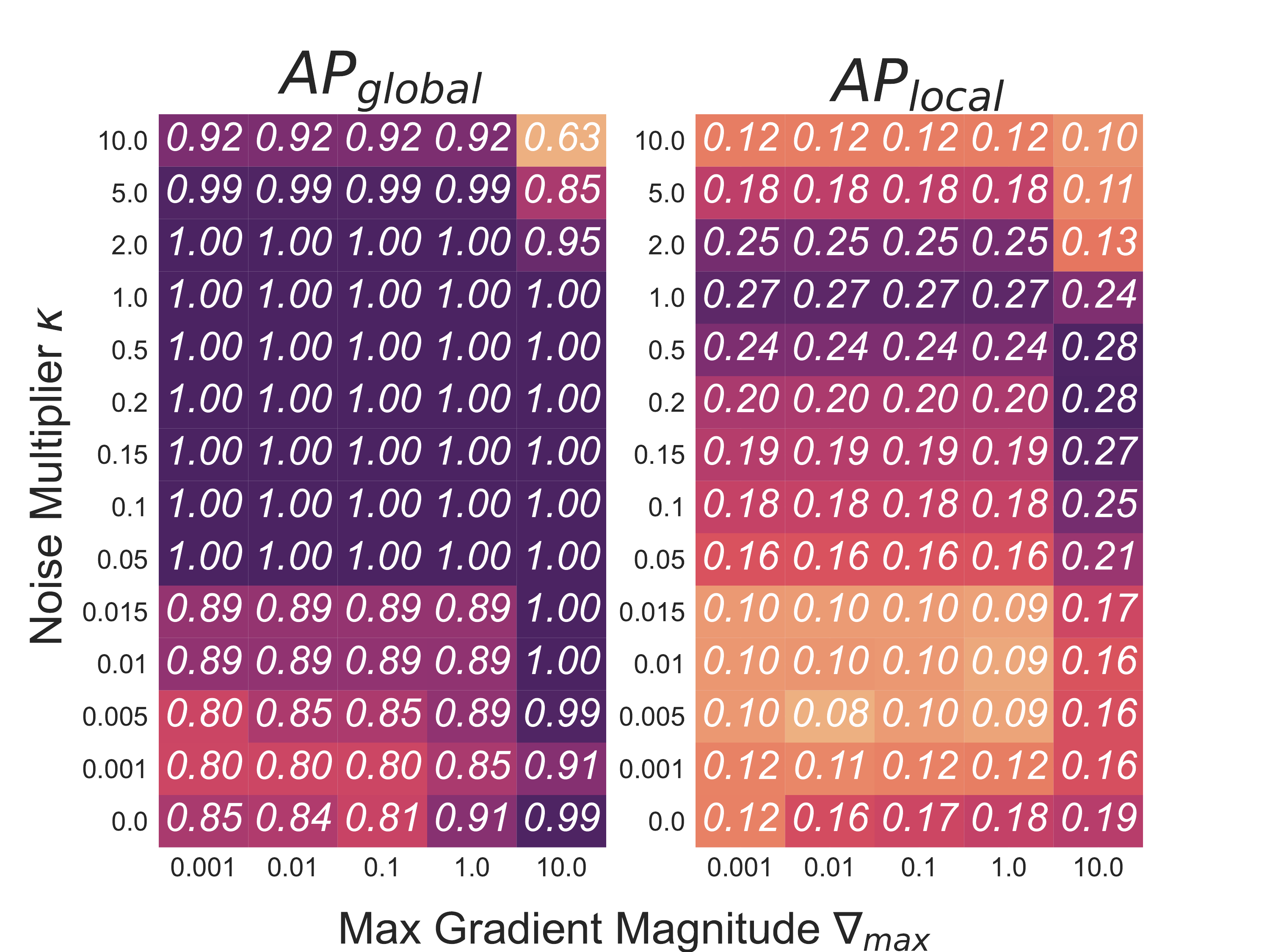}\\
        \vspace{0.5mm}
        \textbf{(a) Philadelphia Payments $\mathcal{D}^{A}$}
    \end{minipage}
    \begin{minipage}[b]{.32\linewidth}
        \center
        \includegraphics[height=4.6cm,trim=0 0 0 0, clip]{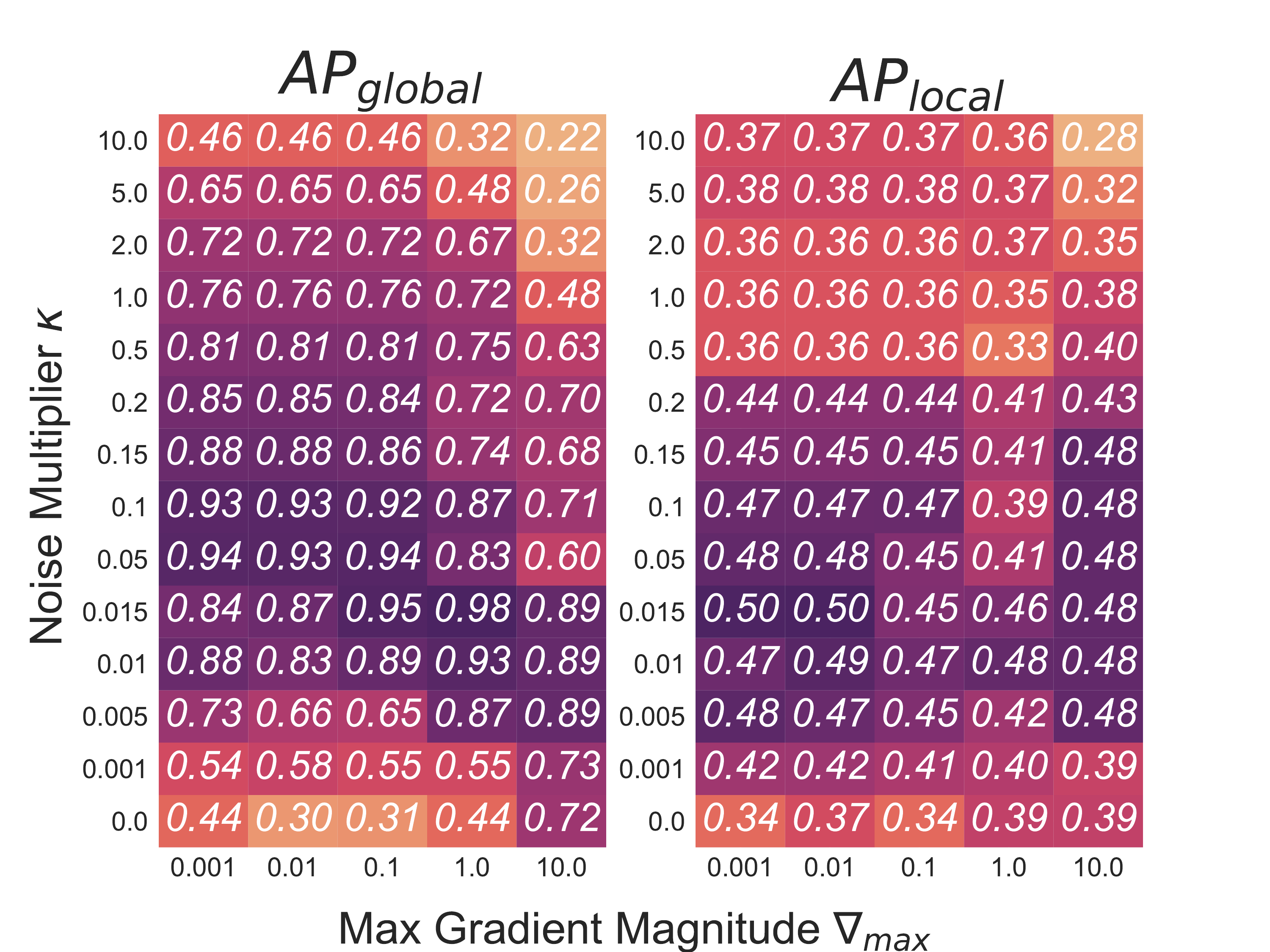}\\
        \vspace{0.5mm}
        \textbf{(b) Chicago Payments $\mathcal{D}^{B}$}
    \end{minipage}
    \begin{minipage}[b]{.32\linewidth}
        \center
        \includegraphics[height=4.6cm,trim=0 0 0 0, clip]{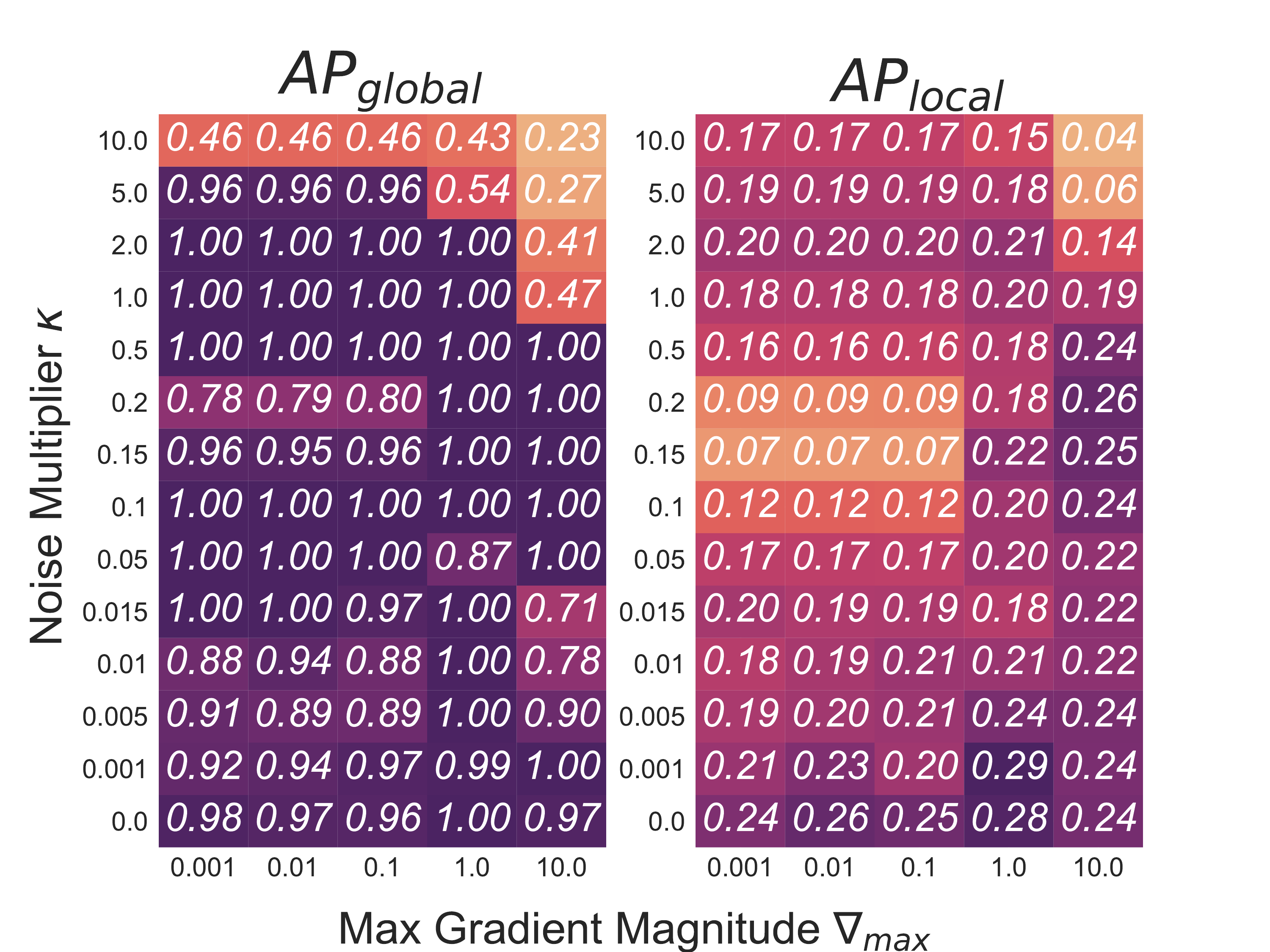}\\
        \vspace{0.5mm}
        \textbf{(c) York Payments $\mathcal{D}^{C}$}
    \end{minipage}
    \vspace{-3mm}
    \caption{Average precision scores obtained in the iid data setting for the global \scalebox{0.9}{$AP_{global}$} and local \scalebox{0.9}{$AP_{local}$} anomalies with \scalebox{0.9}{$\lambda =8$} federated clients when evaluating the grid of distinct DPL configurations over all payment datasets. The scores of a particular \scalebox{0.9}{$\nabla_{max}$} and \scalebox{0.9}{$\kappa$} combination denotes the mean over five distinct public \scalebox{0.9}{$\theta^{*}$} and private \scalebox{0.9}{$\theta'$} model parameter initializations.}
    \label{fig:differential_privacy_results}
    \vspace{-3mm}
\end{figure*}

\section{Experimental Setup}
\label{sec:experimental_setup}

\noindent In this section, we describe the experimental details of the privacy-preserving federated learning setup to audit large-scale JE data.\footnote{Due to the general confidentiality of JE data, we evaluate the proposed methodology based on publicly available real-world datasets to allow for result reproducibility.}

\vspace{0.5mm}

\textbf{Datasets and Data Preparation:} To evaluate the learning capabilities of the learning framework, we use three publicly available datasets of real-world financial payment data. The datasets exhibit high similarity to ERP accounting data, e.g., typical manual payments (SAP T-Code: F-53) or payment runs (SAP T-Code: F-110). 

\begin{itemize}
\item The \textit{City of Philadelphia} payments\footnote{\scalebox{0.9}{\url{https://www.phila.gov/2019-03-29-philadelphias-initial-release-of-city-payments-data/}}} denoted as \scalebox{0.9}{$\mathcal{D}^{A}$} encompass a total of 238,894 payments generated by 58 distinct city departments in 2017. Each payment exhibits $10$ categorical and $1$ numerical attribute(s).

\item The \textit{City of Chicago} payments\footnote{\scalebox{0.9}{\url{https://data.cityofchicago.org/Administration-Finance/Payments/s4vu-giwb/}}} denoted as \scalebox{0.9}{$\mathcal{D}^{B}$} encompass a total of 108,478 payments generated by 54 distinct city departments in 2021. Each payment exhibits $6$ categorical and $1$ numerical attribute(s).

\item The \textit{City of York} payments\footnote{\scalebox{0.9}{\url{https://data.yorkopendata.org/dataset/all-payments-to-suppliers/}}} denoted as \scalebox{0.9}{$\mathcal{D}^{C}$} encompass a total of 102,026 payments generated by 49 distinct city departments in 2020/21 and 2021/22. Each payment exhibits $11$ categorical and $2$ numerical attribute(s).
\end{itemize}

\noindent For all datasets, let \scalebox{0.9}{$\{A_{j}\}_{j=1}^{N}$} denote the set of payments generating city departments. For each department $A_{j}$ the payment's attribute values are pre-processed to derive \scalebox{0.9}{$\hat{x}\in \hat{X}$}. Thereby, the categorical attribute values of \scalebox{0.9}{$x_{i}^{j}$} are converted into \textit{one-hot} numerical tuples of bits \scalebox{0.9}{$\hat{x}_{i}^{j} \in \{0, 1\}^{\upsilon}$}, where $\upsilon$ denotes the number of unique attribute values in \scalebox{0.9}{$x^{j}$}. The numerical attribute values of \scalebox{0.9}{$x_{i}^{l}$} are scaled, according to \scalebox{0.9}{$\hat{x}_{i}^{l} = (x_{i}^{l} - \text{min}(x^{l}) / (\text{max}(x^{l}) - \text{min}(x^{l}))$}, where the $\text{min}$ and $\text{max}$ are obtained over all attribute values in \scalebox{0.9}{$x^{l}$}. 



\vspace{0.5mm}

\textbf{Anomaly Detection Setup:} To establish the anomaly detection setup, we inject a small fraction of 60 (30, 20) randomly sampled global and 140 (70, 30) local anomalies into dataset $\mathcal{D}^{A}$ ($\mathcal{D}^{B}$, $\mathcal{D}^{C}$).\footnote{To create both classes of anomalies, we use the `Faker' project: \\ \scalebox{0.9}{\url{https://github.com/joke2k/faker}}.} We use a symmetrical encoder $q_{\psi}$ and decoder $p_{\phi}$ architecture, as shown in Tab. \ref{tab:architecture}, in all our experiments. In both networks, we apply Leaky-ReLU non-linear activations with scaling factor $\alpha = 0.4$ except in the encoder's bottleneck and decoder's final layer, which comprise Tanh activations. The models are trained for $R=$ 100 (200, 200) communication rounds and $\tau=$ 200 (400, 400) update iterations per round. We use a batch size of $\rho=$ 64 (32, 32) journal entries, apply Adam optimization with $\beta_{1}=0.9$, $\beta_{2}=0.999$, and early stopping once the loss converges. Furthermore, optimize the AEN models using a combined loss-function \citep{schreyer2019a} that computes the reconstruction error of an encoded journal entry $\hat{x}_{i}$, according to: 

\begin{equation}
    \mathcal{L}^{REC}_{\psi, \phi}(\tilde{x}_{i},\hat{x}_{i}) = \vartheta \hspace{1mm} \sum\nolimits_{j=1}^{M} \mathcal{L}^{BCE}_{j, \psi, \phi} + (1 - \vartheta) \hspace{1mm} \sum\nolimits_{l=1}^{K} \mathcal{L}^{MSE}_{l, \psi, \phi} \;,
    \vspace{2mm}
    \label{equ:reconstruction_loss_details}
\end{equation}

\noindent where \scalebox{0.9}{$\tilde{x}_{i} = p_{\phi}(q_{\psi}(\hat{x}_{i}))$} denotes the i-\textit{th} journal entry reconstruction, $M$ the number of categorical attributes, and $K$ the number of numerical attributes. Thereby, \scalebox{0.8}{$\mathcal{L}^{BCE}$} denotes the normalized binary-cross-entropy error of a categorical attribute \scalebox{0.9}{$\hat{x}_{i}^{j}$}, defined as: 

\begin{equation}
    \mathcal{L}^{BCE}_{\psi, \phi}(\tilde{x}_{i}^{j},\hat{x}_{i}^{j}) = \frac{1}{\Upsilon} \sum\nolimits_{\upsilon=1}^{\Upsilon} \tilde{x}_{i,\upsilon}^{j} \log(\hat{x}_{i,\upsilon}^{j}) + (1 - \tilde{x}_{i,\upsilon}^{j}) \log(1 - \hat{x}_{i,\upsilon}^{j}) \;,
    \vspace{2mm}
    \label{equ:categorical_loss_details}
\end{equation}

\noindent where $\upsilon=1, 2, ..., \Upsilon$ corresponds to the number of one-hot encoded attribute dimensions. Furthermore, \scalebox{0.8}{$\mathcal{L}^{MSE}$} denotes the mean-squared error of a given numerical attribute \scalebox{0.9}{$\hat{x}_{i}^{l}$}, defined as: 

\begin{equation}
    \mathcal{L}^{MSE}_{\psi, \phi}(\tilde{x}_{i}^{l},\hat{x}_{i}^{l}) = (\tilde{x}_{i}^{l} - \hat{x}_{i}^{l})^2 \;.
    \vspace{2mm}
    \label{equ:numerical_loss_details}
\end{equation}

\noindent To account for the higher number of categorical attributes in each dataset, we set \scalebox{0.85}{$\vartheta=\frac{2}{3}$} to balance both loss terms.

\begin{table}[t!]
  \vspace*{-2mm}
  \caption{Number of neurons per layer $\ell_{i}$ of the encoder $q_{\psi}$ and decoder $p_{\phi}$ network that constitute the AEN architectures.} 
    \vspace*{-2mm}
  \fontsize{8}{6}\selectfont
  \centering
  \begin{tabular}{l c c c c c c c c}
    \toprule
        \multicolumn{1}{l}{Layer $\ell_{i}$}
        & \multicolumn{1}{c}{$i$ = 1}
        & \multicolumn{1}{c}{2}
        & \multicolumn{1}{c}{3}
        & \multicolumn{1}{c}{4}
        & \multicolumn{1}{c}{5}
        & \multicolumn{1}{c}{6}
        & \multicolumn{1}{c}{7}
        & \multicolumn{1}{c}{8}
        \\
    \midrule
    $q_{\psi}(z|\hat{x})$ & |$\hat{x}$| & 128 & 64 & 32 & 16 & 8 & 4 & 2 \\
    $p_{\phi}(\tilde{x}|z)$ & 2 & 4 & 8 & 16 & 32 & 64 & 128 & |$\hat{x}$| \\
    \bottomrule \\
  \end{tabular}
    \label{tab:architecture}
    \vspace*{-5mm}
\end{table}

\textbf{Federated Learning (FL) Setup:} We establish the FL setup using (i) \textit{non-iid} and (ii) \textit{iid} data partitions for each payment dataset \scalebox{0.9}{$\mathcal{D} \in [\mathcal{D}^{A}, \mathcal{D}^{B}, \mathcal{D}^{C}]$}. In the non-iid setup, the payments of a dataset \scalebox{0.9}{$\mathcal{D}_{i}$} are partitioned according to the dataset's generating city departments \scalebox{0.9}{$\{A_{j}\}_{j=1}^{N}$}. Subsequently, the derived partitions are distributed across the federated clients to yield an approximately equal number of payments per federated client $\lambda$. In the iid setup, the payments are randomly distributed across the federated clients disregarding the generating city department information. In both setups, the payments $x_{i}$ of each client are pre-processed to derive $\hat{x}_{i}$, where \scalebox{0.9}{$\hat{x}_{i} = \{\hat{x}_{i}^{1}, \hat{x}_{i}^{2}, ..., \hat{x}_{i}^{M}; \hat{x}_{i}^{1}, \hat{x}_{i}^{2}, ..., \hat{x}_{i}^{K}\}$}. Finally, global and local anomalies are injected into the payments residing at the first client to serve as the basis for our evaluation. In all our experiments, we partition a dataset into $\gamma=8$ sub-datasets. We use \scalebox{0.9}{$\lambda \in [1, 4, 8]$} federated clients and the corresponding number of data partitions. The number of payments per client is shown in Tab. \ref{tab:payments_per_client}.

\begin{table}[t!]
  \vspace*{-0mm}
  \caption{Number of payments per federated client $\lambda$ used to evaluate the distinct federated learning setups.} 
    \vspace*{-2mm}
  \fontsize{8}{5}\selectfont
  \centering
  \begin{tabular}{l c c c c c c c c}
    \toprule
        \multicolumn{1}{l}{Data}
        & \multicolumn{1}{c}{$\lambda$ = 1}
        & \multicolumn{1}{c}{2}
        & \multicolumn{1}{c}{3}
        & \multicolumn{1}{c}{4}
        & \multicolumn{1}{c}{5}
        & \multicolumn{1}{c}{6}
        & \multicolumn{1}{c}{7}
        & \multicolumn{1}{c}{8}
        \\
    \midrule
    \scalebox{0.85}{$\mathcal{D}^{A}_{niid}$} & \scalebox{0.85}{24,592} & \scalebox{0.85}{22,602} & \scalebox{0.85}{37,540} & \scalebox{0.85}{53,716} & \scalebox{0.85}{28,974} & \scalebox{0.85}{23,706} & \scalebox{0.85}{33,213} & \scalebox{0.85}{14,751}\\ \addlinespace[0.05cm]
    \scalebox{0.85}{$\mathcal{D}^{B}_{niid}$} & \scalebox{0.85}{12,987} & \scalebox{0.85}{12,886} & \scalebox{0.85}{12,884} & \scalebox{0.85}{16,126} & \scalebox{0.85}{12,882} & \scalebox{0.85}{13,903} & \scalebox{0.85}{14,079} & \scalebox{0.85}{12,781}\\ \addlinespace[0.05cm]
    \scalebox{0.85}{$\mathcal{D}^{C}_{niid}$} & \scalebox{0.85}{12,972} & \scalebox{0.85}{13,864} & \scalebox{0.85}{12,641} & \scalebox{0.85}{13,448} & \scalebox{0.85}{14,810} & \scalebox{0.85}{12,651} & \scalebox{0.85}{12,633} & \scalebox{0.85}{9,057}\\
    \midrule
    \scalebox{0.85}{$\mathcal{D}^{A}_{iid}$} & \scalebox{0.85}{14,266} & \scalebox{0.85}{45,656} & \scalebox{0.85}{29,002} & \scalebox{0.85}{30,722} & \scalebox{0.85}{24,472} & \scalebox{0.85}{35,252} & \scalebox{0.85}{8,287} & \scalebox{0.85}{51,437}\\ \addlinespace[0.05cm]
    \scalebox{0.85}{$\mathcal{D}^{B}_{iid}$} & \scalebox{0.85}{6,489} & \scalebox{0.85}{20,829} & \scalebox{0.85}{12,700} & \scalebox{0.85}{14,420} & \scalebox{0.85}{8,170} & \scalebox{0.85}{18,950} & \scalebox{0.85}{8,369} & \scalebox{0.85}{18,751}\\ \addlinespace[0.05cm]
    \scalebox{0.85}{$\mathcal{D}^{C}_{iid}$} & \scalebox{0.85}{5,483} & \scalebox{0.85}{20,023} & \scalebox{0.85}{11,893} & \scalebox{0.85}{13,613} & \scalebox{0.85}{7,363} & \scalebox{0.85}{18,143} & \scalebox{0.85}{7,563} & \scalebox{0.85}{17,945}\\
    \bottomrule \\
  \end{tabular}
    \label{tab:payments_per_client}
    \vspace*{-7mm}
\end{table}

\vspace{0.5mm}

\textbf{Split Learning (SL) Setup:} To facilitate the SL setup we set the cut-layer to \scalebox{0.8}{$\ell_{c}^{q}$} = 1 and \scalebox{0.8}{$\ell_{c}^{p}$} = 1. Thereby, for each decentral client model \scalebox{0.9}{$\theta_{i}^{\omega}$} the parameters of the first encoder (last decoder) layer are kept private and are not transferred for a central model aggregation. We implement the FL/SL setup by adapting the Flower library \cite{beutel2020}. 

\vspace{0.5mm}

\textbf{Differential Private Learning (DPL) Setup:} To evaluate the DPL setup we run several experiment using maximum norm thresholds \scalebox{0.9}{$\nabla_{max} \in [0.001, 0.01, 0.1, 1.0, 10.0]$}. We applied `flat' gradient clipping that calculates the norm of the gradient over all parameters. Furthermore, we evaluate a set of noise multiplier configurations \scalebox{0.9}{$\kappa \in [0.0, 0.001, 0.005, 0.01, 0.015, 0.1, 0.15, 0.2]$}. To implement the DP setup we use the Opacus library \cite{yousefpour2021}.

\vspace{0.5mm}

\textbf{Evaluation Metric:} To quantitatively assess the anomaly detection capability of the fine-tuned models we derive the average precision $AP$ over the sorted payments reconstruction errors \scalebox{0.8}{$\mathcal{L}^{REC}_{\psi, \phi}$}. The $AP$ summarizes the precision-recall curve, formally defined by:  

\begin{equation}
    AP(\mathcal{L}^{REC}_{\psi, \phi}) = \sum\nolimits_{i=1}^{N} (R_{i}- R_{i-1}) P_{i},
    \vspace{2mm}
    \label{equ:average_precision}
\end{equation}

\noindent where \scalebox{0.9}{$P_{i}($}\scalebox{0.75}{$\mathcal{L}^{REC}_{\psi, \phi}$}\scalebox{0.9}{$) = TP/(TP + FP)$} denotes the detection precision, and \scalebox{0.9}{$R_{i}($}\scalebox{0.75}{$\mathcal{L}^{REC}_{\psi, \phi}$}\scalebox{0.9}{$)=TP /(TP + FN)$} denotes the detection recall of the i-\textit{th} reconstruction error threshold.

\section{Experimental Results}
\label{sec:experimental_results}

In this section, we present and assess the results of three Research Questions (RQs) when evaluating the different learning setups that constitute the proposed framework.

\begin{table}[ht!]
\vspace*{-0mm}
\caption{Average precision scores obtained for all journal entry anomalies $AP_{all}$, global anomalies $AP_{global}$, and local anomalies $AP_{local}$ with different number of clients \scalebox{0.9}{$\lambda \in [1, 4, 8]$} and non-iid (\scalebox{0.9}{$\mathcal{D}_{niid}$}) and iid (\scalebox{0.9}{$\mathcal{D}_{iid}$}) data configurations.}
\vspace*{-2mm}
\fontsize{8}{7}\selectfont
\centering
\begin{tabular}{l l c | c | c | c }
\toprule
    \multicolumn{1}{l}{Model}
    & \multicolumn{1}{l}{Data}
    & \multicolumn{1}{l}{$\lambda$}
    & \multicolumn{1}{c}{$AP_{all} \uparrow$}
    & \multicolumn{1}{c}{$AP_{global} \uparrow$}
    & \multicolumn{1}{c}{$AP_{local} \uparrow$}
    \\

\midrule
 & \scalebox{0.8}{} & 1 & 0.511 $\pm$ \scalebox{0.85}{0.15} & 0.693 $\pm$ \scalebox{0.85}{0.08} & 0.153 $\pm$ \scalebox{0.85}{0.09}\\
AEN\scalebox{0.7}{\textit{FL}} & \scalebox{0.85}{$\mathcal{D}^{A}_{niid}$} & 4 & 0.600 $\pm$ \scalebox{0.85}{0.15} & 0.831 $\pm$ \scalebox{0.85}{0.14} & 0.206 $\pm$ \scalebox{0.85}{0.12}\\
 & \scalebox{0.8}{} & 8 & 0.645 $\pm$ \scalebox{0.85}{0.09} & 0.789 $\pm$ \scalebox{0.85}{0.02} & 0.238 $\pm$ \scalebox{0.85}{0.06}\\
\midrule
 & \scalebox{0.8}{} & 1 & 0.622 $\pm$ \scalebox{0.85}{0.04} & 0.993 $\pm$ \scalebox{0.85}{0.01} & 0.180 $\pm$ \scalebox{0.85}{0.03}\\
AEN\scalebox{0.7}{\textit{FL}} & \scalebox{0.85}{$\mathcal{D}^{A}_{iid}$} & 4 & 0.618 $\pm$ \scalebox{0.85}{0.06} & 0.936 $\pm$ \scalebox{0.85}{0.04} & 0.178 $\pm$ \scalebox{0.85}{0.04}\\
 & \scalebox{0.8}{} & 8 & \textbf{0.619} $\pm$ \scalebox{0.85}{\textbf{0.06}} & \textbf{1.000} $\pm$ \scalebox{0.85}{\textbf{0.01}} & 0.178 $\pm$ \scalebox{0.85}{0.04}\\
\midrule
AEN\scalebox{0.7}{\textit{Base}} & \scalebox{0.85}{$\mathcal{D}^{A}$} & - & 0.530 $\pm$ \scalebox{0.85}{0.10} & 0.479 $\pm$ \scalebox{0.85}{0.17} & \textbf{0.232} $\pm$ \scalebox{0.85}{\textbf{0.06}} \vspace{0.4mm} \\ 
\hline \hline & \\[\dimexpr-\normalbaselineskip+2pt]
\addlinespace[0.1cm]
 & \scalebox{0.8}{} & 1 & 0.595 $\pm$ \scalebox{0.85}{0.15} & 0.498 $\pm$ \scalebox{0.85}{0.08} & 0.254 $\pm$ \scalebox{0.85}{0.09}\\
AEN\scalebox{0.7}{\textit{FL}} & \scalebox{0.85}{$\mathcal{D}^{B}_{niid}$} & 4 & 0.675 $\pm$ \scalebox{0.85}{0.10} & 0.624 $\pm$ \scalebox{0.85}{0.11} & 0.257 $\pm$ \scalebox{0.85}{0.08}\\
 & \scalebox{0.8}{} & 8 & 0.668 $\pm$ \scalebox{0.85}{0.13} & 0.646 $\pm$ \scalebox{0.85}{0.16} & 0.259 $\pm$ \scalebox{0.85}{0.08}\\
\midrule
 & \scalebox{0.8}{} & 1 & 0.622 $\pm$ \scalebox{0.85}{0.35} & 0.565 $\pm$ \scalebox{0.85}{0.34} & 0.306 $\pm$ \scalebox{0.85}{0.17}\\
AEN\scalebox{0.7}{\textit{FL}} & \scalebox{0.85}{$\mathcal{D}^{B}_{iid}$} & 4 & 0.800 $\pm$ \scalebox{0.85}{0.34} & 0.812 $\pm$ \scalebox{0.85}{019} & 0.365 $\pm$ \scalebox{0.85}{0.10}\\
 & \scalebox{0.8}{} & 8 & 0.859 $\pm$ \scalebox{0.85}{0.18} & \textbf{0.870} $\pm$ \scalebox{0.85}{\textbf{0.15}} & 0.394 $\pm$ \scalebox{0.85}{0.08}\\
\midrule
AEN\scalebox{0.7}{\textit{Base}} & \scalebox{0.85}{$\mathcal{D}^{B}$} & - & \textbf{0.869} $\pm$ \scalebox{0.85}{\textbf{0.14}} & 0.375 $\pm$ \scalebox{0.85}{0.07} & \textbf{0.574} $\pm$ \scalebox{0.85}{\textbf{0.04}} \vspace{0.4mm} \\ 
\hline \hline & \\[\dimexpr-\normalbaselineskip+2pt]
\addlinespace[0.1cm]
 & \scalebox{0.8}{} & 1 & 0.536 $\pm$ \scalebox{0.85}{0.07} & 0.715 $\pm$ \scalebox{0.85}{0.17} & 0.149 $\pm$ \scalebox{0.85}{0.05}\\
AEN\scalebox{0.7}{\textit{FL}} & \scalebox{0.9}{$\mathcal{D}^{C}_{niid}$} & 4 & 0.628 $\pm$ \scalebox{0.85}{0.12} & 0.875 $\pm$ \scalebox{0.85}{0.14} & 0.164 $\pm$ \scalebox{0.85}{0.04}\\
 & \scalebox{0.8}{} & 8 & 0.646 $\pm$ \scalebox{0.85}{0.04} & 0.939 $\pm$ \scalebox{0.85}{0.08} & 0.159 $\pm$ \scalebox{0.85}{0.03}\\
\midrule
 & \scalebox{0.8}{} & 1 & 0.580 $\pm$ \scalebox{0.85}{0.01} & 0.807 $\pm$ \scalebox{0.85}{0.14} & 0.103 $\pm$ \scalebox{0.85}{0.01}\\
AEN\scalebox{0.7}{\textit{FL}} & \scalebox{0.85}{$\mathcal{D}^{C}_{iid}$} & 4 & 0.739 $\pm$ \scalebox{0.85}{0.15} & 0.934 $\pm$ \scalebox{0.85}{0.15} & 0.202 $\pm$ \scalebox{0.85}{0.06}\\
 & \scalebox{0.8}{} & 8 & \textbf{0.749} $\pm$ \scalebox{0.85}{\textbf{0.03}} & \textbf{1.000} $\pm$ \scalebox{0.85}{\textbf{0.01}} & 0.207 $\pm$ \scalebox{0.85}{0.01}\\
\midrule
AEN\scalebox{0.7}{\textit{Base}} & \scalebox{0.85}{$\mathcal{D}^{C}$} & - & 0.504 $\pm$ \scalebox{0.85}{0.07} & 0.493 $\pm$ \scalebox{0.85}{0.16} & \textbf{0.219} $\pm$ \scalebox{0.85}{\textbf{0.08}}\\ 
\bottomrule \\ \addlinespace[-0.15cm]
\multicolumn{6}{l}{\scalebox{0.8}{*Variances originate from training using five distinct random seeds.}}
\end{tabular}
\label{tab:anomaly_scores}
\vspace*{-0.9cm}
\end{table}

\vspace{0.5mm}

\begin{description}
\item[RQ 1:] \textit{Will the accounting anomaly detection performance benefit from applying different FL setups?}
\end{description}

\vspace{0.5mm}

\noindent We evaluate the FL setup by comparing the learned model's anomaly detection capabilities to baseline models \scalebox{0.9}{$AEN_{Base}$}. For the dataset, \scalebox{0.9}{$\mathcal{D}^{A}$} (\scalebox{0.9}{$\mathcal{D}^{B}, \mathcal{D}^{C}$}) the baseline is trained for 100 (200, 200) epochs and 200 (400, 400) iterations per epoch on the entire dataset as proposed in \citep{schreyer2017}. The distinct FL models \scalebox{0.9}{$AEN_{FL}$} are trained with \scalebox{0.9}{$\nabla_{max} = 1.0$}, \scalebox{0.9}{$\kappa = 0.1$}, and \scalebox{0.8}{$\ell_{c}^{q}$} = \scalebox{0.8}{$\ell_{c}^{p}$} = 1 (disregarding DPL and SL variations) in the \textit{non-iid} (\scalebox{0.9}{$\mathcal{D}_{niid}$}) and \textit{iid} (\scalebox{0.9}{$\mathcal{D}_{iid}$}) setting to optimize the learning objective defined in Eq. \ref{equ:reconstruction_loss_details}.

\vspace{0.5mm}

\textbf{Results.} Table \ref{tab:anomaly_scores} shows the average precision obtained for different client configurations \scalebox{0.9}{$\lambda$}, both anomaly classes \scalebox{0.9}{$AP_{all}$}, the global anomalies \scalebox{0.9}{$AP_{global}$}, and the local anomalies \scalebox{0.9}{$AP_{local}$}. For all datasets, increasing the number of clients \scalebox{0.9}{$\lambda > 1$} yields an improved detection performance of the FL models. This observation holds for both anomaly classes. When comparing the FL models \scalebox{0.9}{$AEN_{FL}$} trained on \scalebox{0.9}{$\lambda \geq 4$} clients to the baseline models \scalebox{0.9}{$AEN_{Base}$}, the FL models outperform the baselines in detecting global anomalies. This performance gap can be attributed to two effects of the different learning setups. First, the FL setup yields a high global anomaly detection performance in the non-iid and iid settings since such anomalies violate general posting patterns observable at multiple clients. Second, the baseline setup yields a low global anomaly detection performance since such anomalies tend to be learned due to the comparably higher model capacity (resulting in a low reconstruction error \scalebox{0.9}{$\mathcal{L}^{REC}$}). In summary, the results provide initial evidence of the benefits of learning specialized accounting `meta'-models from multiple clients, encouraging the learning of industry, sector, or jurisdiction-specific audit models.

\begin{figure}[t!]
    \begin{minipage}[b]{1.0\linewidth}
    \center
        \begin{minipage}[b]{0.48\linewidth}
            \center
            \includegraphics[width=4.0cm, angle=0, trim={0.0cm 0.0cm 0.0cm 0.0cm}]{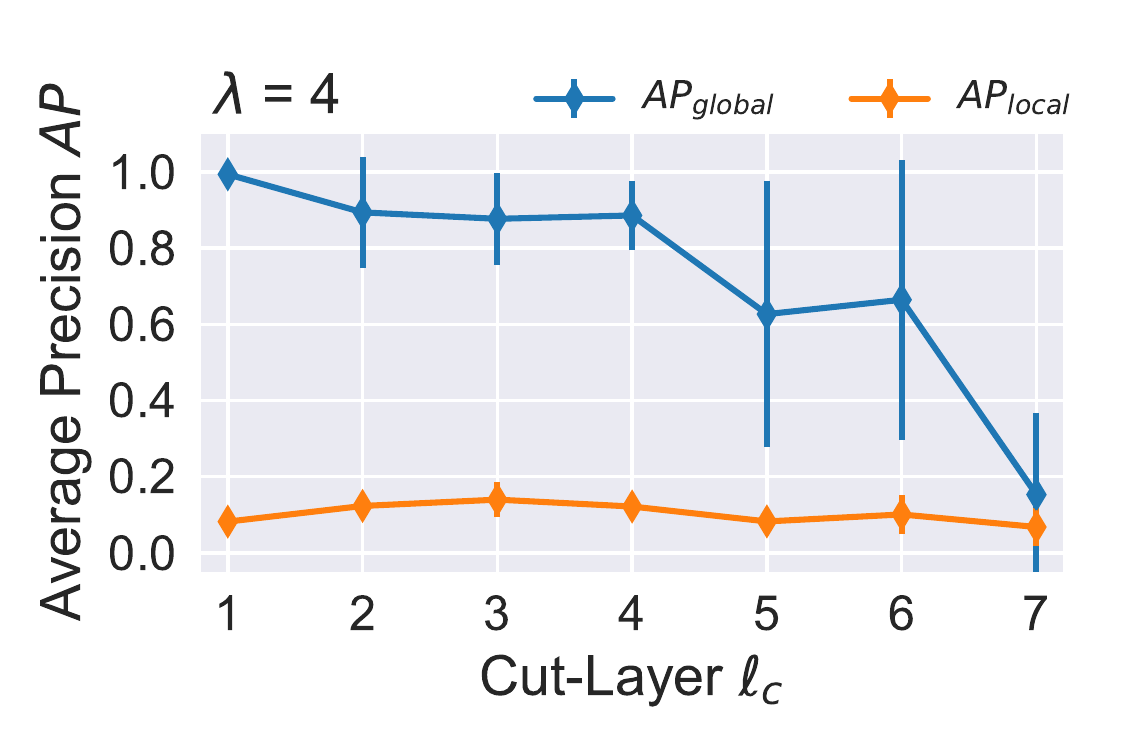}\\
	    \end{minipage}
	   \begin{minipage}[b]{0.48\linewidth}
            \center
            \includegraphics[width=4.0cm, angle=0, trim={0.0cm 0.0cm 0.0cm 0.0cm}]{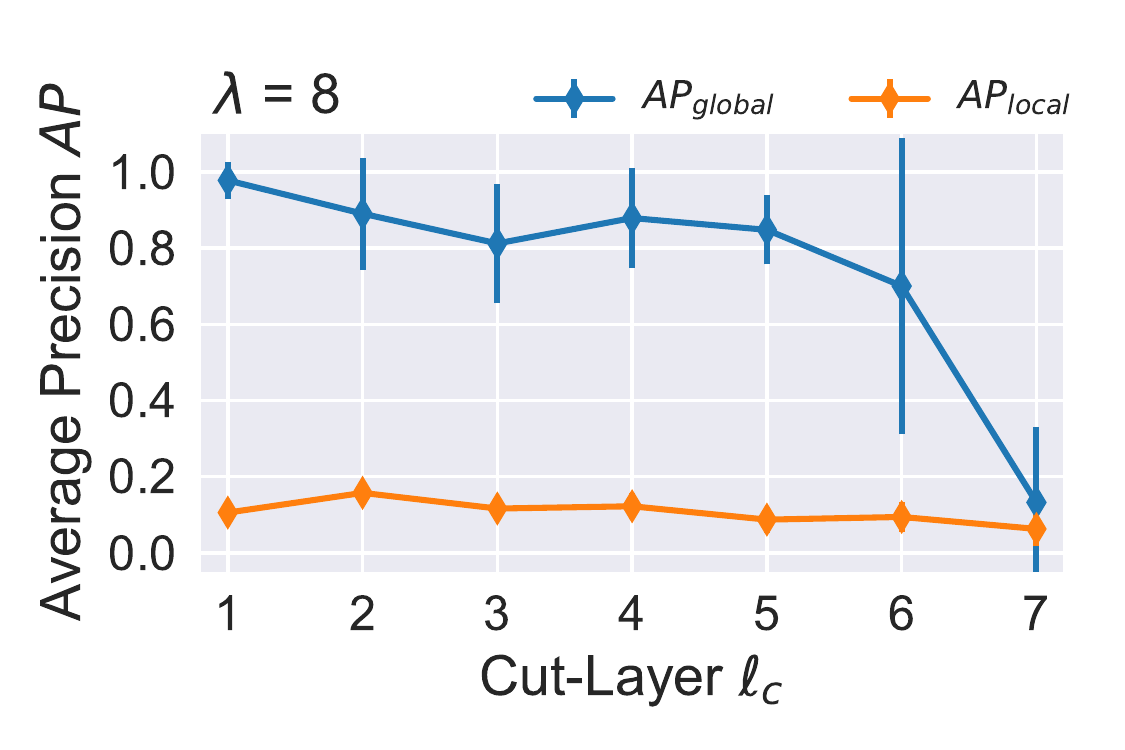}\\
	    \end{minipage}
	\vspace{-0.5mm}
	\end{minipage}
    \textbf{(a) Philadelphia City Payments $\mathcal{D}^{A}$}\\
    \vspace{0.5mm}
	
	\begin{minipage}[b]{1.0\linewidth}
    \center
        \begin{minipage}[b]{0.48\linewidth}
            \center
            \includegraphics[width=4.0cm, angle=0, trim={0.0cm 0.0cm 0.0cm 0.0cm}]{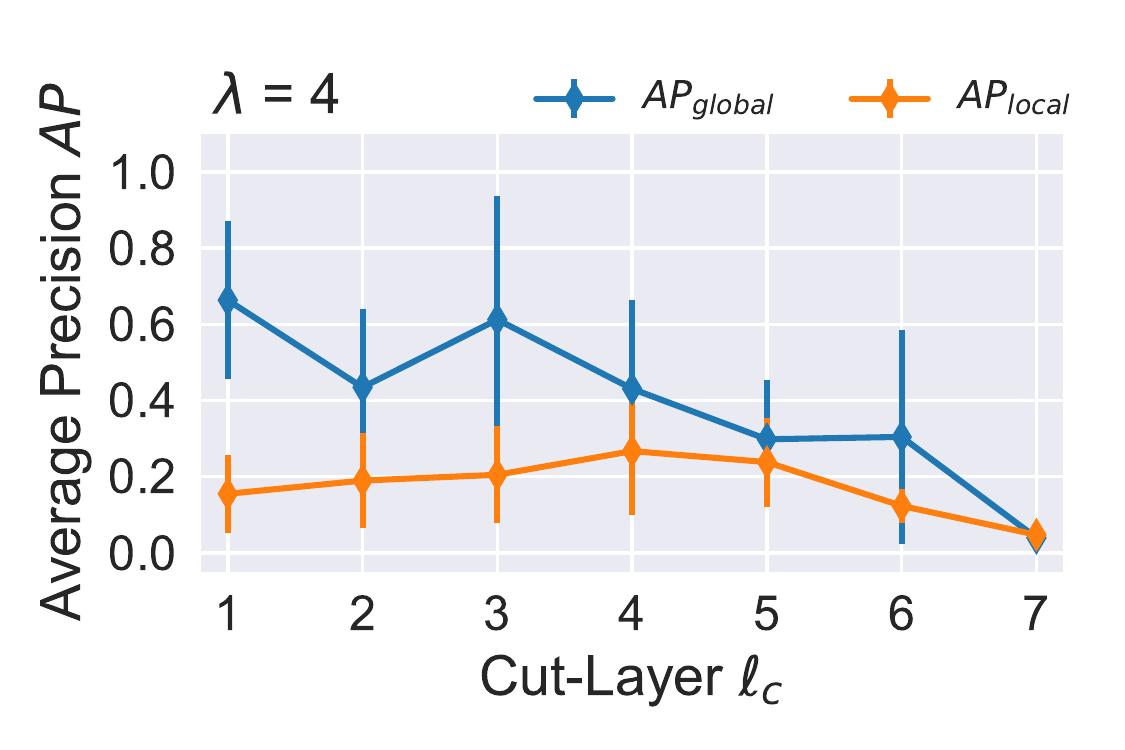}\\
	    \end{minipage}
	   \begin{minipage}[b]{0.48\linewidth}
            \center
            \includegraphics[width=4.0cm, angle=0, trim={0.0cm 0.0cm 0.0cm 0.0cm}]{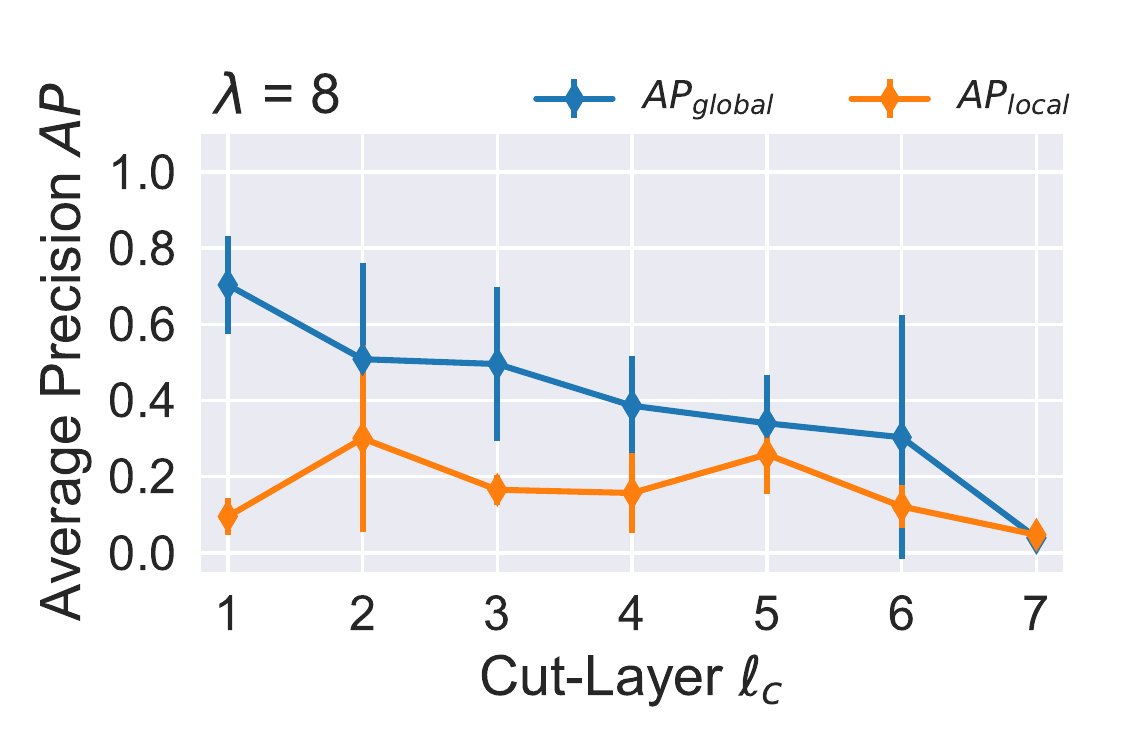}\\
	    \end{minipage}
	\vspace{-0.5mm}
	\end{minipage}
    \textbf{(b) Chicago City Payments $\mathcal{D}^{B}$}\\
    \vspace{0.5mm}
	
	\begin{minipage}[b]{1.0\linewidth}
    \center
        \begin{minipage}[b]{0.48\linewidth}
            \center
            \includegraphics[width=3.9cm, angle=0, trim={0.0cm 0.0cm 0.0cm 0.0cm}]{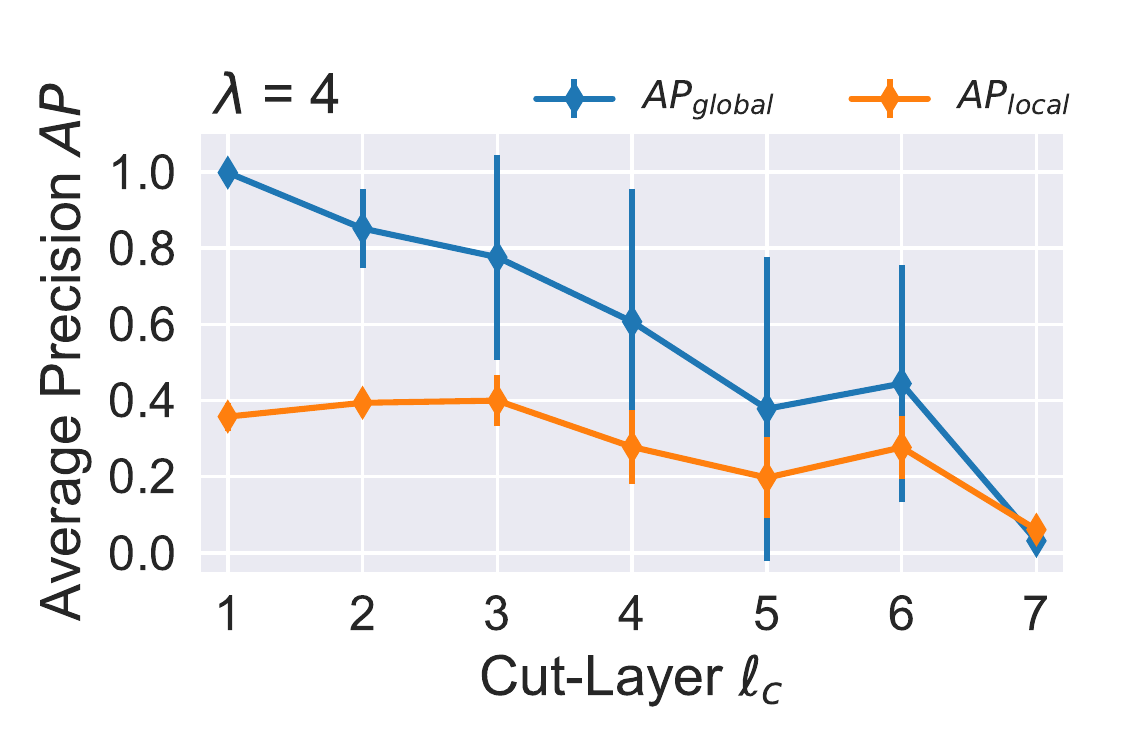}\\
	    \end{minipage}
	   \begin{minipage}[b]{0.48\linewidth}
            \center
            \includegraphics[width=3.9cm, angle=0, trim={0.0cm 0.0cm 0.0cm 0.0cm}]{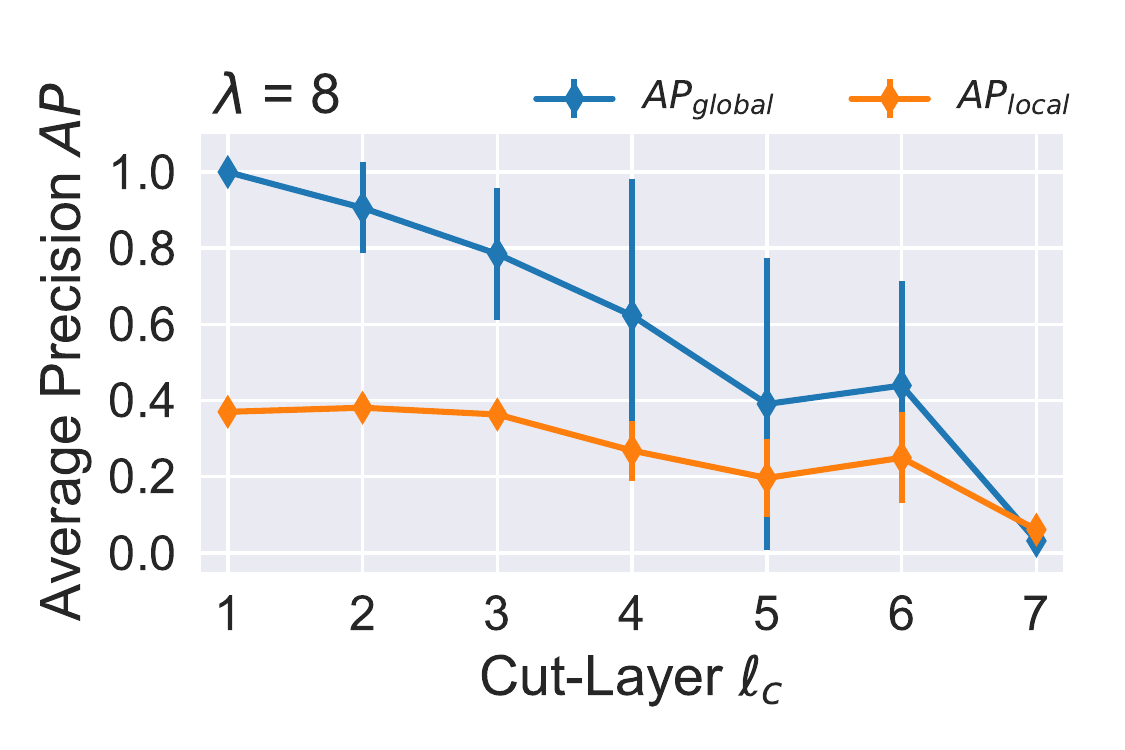}\\
	    \end{minipage}
	\vspace{-0.5mm}
	\end{minipage}
    \textbf{(c) York City Payments $\mathcal{D}^{C}$}\\
    \vspace{-2mm}
    
    \caption{Average precision scores obtained for global anomalies $AP_{global}$, and local anomalies $AP_{local}$ (variances originate from five distinct seed initializations) over different cut-layer configurations \scalebox{0.9}{$\ell_{c} \in [1, 2, ..., 7]$} and city payment dataset.}
	\label{fig:vis_split_learning_setup}
	\vspace{-0.5cm}
\end{figure}

\begin{description}
\item[RQ 2:] \textit{Will the accounting anomaly detection performance be impacted by applying different DPL setups?}
\end{description}

\vspace{0.5mm}

We evaluate the anomaly detection performance of the federated learned AEN models over different DPL parameter configurations \scalebox{0.9}{$\kappa$} and \scalebox{0.9}{$\nabla_{max}$}. The distinct FL models \scalebox{0.9}{$AEN_{FL}$} are trained using \scalebox{0.9}{$\lambda$=8} federated clients, and \scalebox{0.8}{$\ell_{c}^{q}$} = \scalebox{0.8}{$\ell_{c}^{p}$} = 1 (disregarding FL and SL variations) to optimize the learning objective defined in Eq. \ref{equ:reconstruction_loss_details}. 

\vspace{0.5mm}

\textbf{Results.} Figure \ref{fig:differential_privacy_results} shows the average precision \scalebox{0.9}{$AP_{global}$} and \scalebox{0.9}{$AP_{local}$} given the distinct DP parameter configurations. For all datasets, a gradient magnitude threshold \scalebox{0.9}{$\nabla_{max} \geq 1.0$} allows for a high anomaly detection performance of both anomaly classes. Similar detection precisions are achievable by \scalebox{0.9}{$\nabla_{max} \leq 1.0$} reducing the AEN models vulnerability to data leakage attacks. Furthermore, increasing the amount of added noise up to \scalebox{0.9}{$\kappa \leq 1.0$} improves the AEN's anomaly detection capability of the models for both anomaly classes. This potentially counterintuitive observation can be attributed to the increased robustness of learned payment representations. The positive effect of deliberately added noise was also shown by Vincent et al. \cite{vincent2008} in their work on \textit{Denoising AENs}. However, when increasing the amount of added noise \scalebox{0.9}{$\kappa \geq 10.0$}, the detection performance starts to decline again. In summary, the observed model performances indicate the applicability of DPL in an audit context while mitigating the leakage of confidential client data.

\vspace{0.5mm}

\begin{description}
\item[RQ 3:] \textit{Will the accounting anomaly detection performance decrease when applying different Split Learning setups?}
\end{description}

\vspace{0.5mm}

\noindent We evaluate the SL setup by comparing the learned FL model's anomaly detection capabilities over distinct cut-layer \scalebox{0.9}{$\ell_{c} \in [1, 2, ..., 7]$} configurations. The \scalebox{0.9}{$AEN_{FL}$} are trained with \scalebox{0.9}{$\nabla_{max} = 100$}, \scalebox{0.9}{$\kappa = 0.0$}, and \scalebox{0.9}{$\lambda \in [4, 8]$} federated clients (disregarding DPL setup variations).

\vspace{0.5mm}

\textbf{Results.} Figure \ref{fig:vis_split_learning_setup} shows the average precision scores obtained for global anomalies \scalebox{0.9}{$AP_{global}$}, and local anomalies \scalebox{0.9}{$AP_{local}$} given the cut-layer configurations \scalebox{0.9}{$\ell_{c}$}. The obtained results demonstrate that reducing the number of shared public model parameters \scalebox{0.9}{$\theta^{\xi} =\{\psi^{\xi} \cup \, \phi^{\xi}\}$} decreases the anomaly detection performance. This observation holds in particular for the global anomalies and might not come as a surprise since such anomalies often violate general accounting principles observable at multiple clients. Therefore, the detection of global anomalies unusual for a specific industry, sector, or jurisdiction constitutes the idea of knowledge sharing. In summary, it becomes evident that audit firms can adjust the degree of parameter sharing based on their client's consent to detect industry, sector, or jurisdiction-specific global anomalies.

\section{Summary}
\label{sec:summary}

In this work, we proposed a federated learning framework to learn audit models in a privacy-preserving manner from real-world accounting data. Based on three deliberately designed experimental setups, we demonstrated the framework's ability to detect anomalies using knowledge learned from multiple clients. The experimental results provide initial evidence that such a learning setup offers a promising avenue for the learning `meta' audit models to provide enhanced assurance services. Especially, when auditing a particular industry, sector, or jurisdiction. We believe that the future analysis of such models and learned representations will provide novel insights for audit firms that benefit their clients without sharing confidential data.


\bibliographystyle{abbrv}
\bibliography{library}

\begin{thebibliography}{10}

\bibitem{AICPAEthics2014}
{\em AICPA Code of Professional Conduct}.
\newblock American Institute of Certified Public Accountants (AICPA), 2014.

\bibitem{abadi2016}
M.~Abadi, A.~Chu, I.~Goodfellow, H.~B. McMahan, I.~Mironov, K.~Talwar, and
  L.~Zhang.
\newblock {Deep Learning with Differential Privacy}.
\newblock In {\em {ACM SIGSAC Conference on Computer and Communications
  Security}}, pages 308--318, 2016.

\bibitem{alles2015}
M.~G. Alles.
\newblock {Drivers of the Use and Facilitators and Obstacles of the Evolution
  of Big Data by the Audit Profession}.
\newblock {\em Accounting Horizons}, 29(2):439--449, 2015.

\bibitem{appelbaum2016}
D.~Appelbaum.
\newblock {Securing Big Data Provenance for Auditors: The Big Data Provenance
  Black Box as Reliable Evidence}.
\newblock {\em Journal of Emerging Technologies in Accounting}, 13(1):17--36,
  2016.

\bibitem{Argyrou2012}
A.~Argyrou.
\newblock {Auditing Journal Entries Using Self-Organizing Map}.
\newblock In {\em Proceedings of the Eighteenth Americas Conference on
  Information Systems}, 2012.

\bibitem{Bay2002}
S.~Bay, K.~Kumaraswamy, M.~G. Anderle, R.~Kumar, D.~M. Steier, A.~Blvd, and
  S.~Jose.
\newblock {Large Scale Detection of Irregularities in Accounting Data}.
\newblock In {\em Sixth International Conference on Data Mining}, pages 75--86.
  IEEE, 2006.

\bibitem{bengio2013}
Y.~Bengio, A.~Courville, and P.~Vincent.
\newblock Representation learning: A review and new perspectives.
\newblock {\em IEEE transactions on pattern analysis and machine intelligence},
  35(8):1798--1828, 2013.

\bibitem{beutel2020}
D.~J. Beutel, T.~Topal, A.~Mathur, X.~Qiu, T.~Parcollet, P.~P. de~Gusm{\~a}o,
  and N.~D. Lane.
\newblock Flower: A friendly federated learning research framework.
\newblock {\em arXiv preprint arXiv:2007.14390}, 2020.

\bibitem{Breunig2000}
M.~M. Breunig, H.-P. Kriegel, R.~T. Ng, and J.~Sander.
\newblock {LOF: Identifying Density-based Local Outliers}.
\newblock In {\em Proceedings of the 2000 ACM SIGMOD international conference
  on Management of data}, pages 93--104, 2000.

\bibitem{carlini2020}
N.~Carlini, F.~Tramer, E.~Wallace, M.~Jagielski, A.~Herbert-Voss, K.~Lee,
  A.~Roberts, T.~Brown, D.~Song, U.~Erlingsson, et~al.
\newblock {Extracting Training Data from Large Language Models}.
\newblock {\em arXiv preprint arXiv:2012.07805}, 2020.

\bibitem{chan2004}
D.~Chan, A.~Ferguson, D.~Simunic, and D.~Stokes.
\newblock A spatial analysis and test of oligopolistic competition in the
  market for audit services.
\newblock Technical report, Working paper, University of British Columbia,
  2004.

\bibitem{cho2020}
S.~Cho, M.~A. Vasarhelyi, T.~Sun, and C.~Zhang.
\newblock Learning from machine learning in accounting and assurance.
\newblock {\em Journal of Emerging Technologies in Accounting}, 17(1):1--10,
  2020.

\bibitem{dankar2012}
F.~K. Dankar and K.~El~Emam.
\newblock The application of differential privacy to health data.
\newblock In {\em Proceedings of the 2012 Joint EDBT/ICDT Workshops}, pages
  158--166, 2012.

\bibitem{dickey2019}
G.~Dickey, S.~Blanke, and L.~Seaton.
\newblock Machine learning in auditing.
\newblock {\em The CPA Journal}, pages 16--21, 2019.

\bibitem{dwork2008}
C.~Dwork.
\newblock {Differential Privacy: A Survey of Results}.
\newblock In {\em {International Conference on Theory and Applications of
  Models of Computation}}, pages 1--19. Springer, 2008.

\bibitem{dwork2006a}
C.~Dwork, K.~Kenthapadi, F.~McSherry, I.~Mironov, and M.~Naor.
\newblock {Our data, ourselves: Privacy via Distributed Noise Generation}.
\newblock In {\em Annual International Conference on the Theory and
  Applications of Cryptographic Techniques}, pages 486--503. Springer, 2006.

\bibitem{dwork2014}
C.~Dwork, A.~Roth, et~al.
\newblock {The Algorithmic Foundations of Differential Privacy}.
\newblock {\em {Foundations and Trends in Theoretical Computer Science}},
  9(3-4):211--407, 2014.

\bibitem{fukuchi2017}
K.~Fukuchi, Q.~K. Tran, and J.~Sakuma.
\newblock Differentially private empirical risk minimization with input
  perturbation.
\newblock In {\em International Conference on Discovery Science}, pages 82--90.
  Springer, 2017.

\bibitem{grabski2011}
S.~V. Grabski, S.~A. Leech, and P.~J. Schmidt.
\newblock A review of erp research: A future agenda for accounting information
  systems.
\newblock {\em Journal of information systems}, 25(1):37--78, 2011.

\bibitem{gupta2018}
O.~Gupta and R.~Raskar.
\newblock Distributed learning of deep neural network over multiple agents.
\newblock {\em Journal of Network and Computer Applications}, 116:1--8, 2018.

\bibitem{guy2002}
D.~M. Guy, D.~R. Carmichael, and L.~A. Lach.
\newblock {\em Wiley Practitioner's Guide to GAAS 2003: Covering all SASs,
  SSAEs, SSARSs, and Interpretations}.
\newblock Wiley, 2002.

\bibitem{hard2018}
A.~Hard, K.~Rao, R.~Mathews, S.~Ramaswamy, F.~Beaufays, S.~Augenstein,
  H.~Eichner, C.~Kiddon, and D.~Ramage.
\newblock Federated learning for mobile keyboard prediction.
\newblock {\em arXiv preprint arXiv:1811.03604}, 2018.

\bibitem{hawkins2002}
S.~Hawkins, H.~He, G.~Williams, and R.~Baxter.
\newblock {Outlier Detection using Replicator Neural Networks}.
\newblock In {\em International Conference on Data Warehousing and Knowledge
  Discovery}, pages 170--180. Springer, 2002.

\bibitem{hinton2006}
G.~E. Hinton and R.~R. Salakhutdinov.
\newblock {Reducing the Dimensionality of Data with Neural Networks}.
\newblock {\em {Science}}, 313(5786):504--507, 2006.

\bibitem{hogan1999}
C.~E. Hogan and D.~C. Jeter.
\newblock Industry specialization by auditors.
\newblock {\em Auditing: A Journal of Practice \& Theory}, 18(1):1--17, 1999.

\bibitem{hoitash2006}
R.~Hoitash, A.~Kogan, and M.~A. Vasarhelyi.
\newblock Peer-based approach for analytical procedures.
\newblock {\em Auditing: A Journal of Practice \& Theory}, 25(2):53--84, 2006.

\bibitem{ISA240}
IFAC.
\newblock {\em {International Standards on Auditing (ISA) 240: The Auditor's
  Responsibilities Relating to Fraud in an Audit of Financial Statements}}.
\newblock {International Federation of Accountants (IFAC)}, 2009.

\bibitem{Jans2011}
M.~Jans, J.~M. Van Der~Werf, N.~Lybaert, and K.~Vanhoof.
\newblock {A Business Process Mining Application for Internal Transaction Fraud
  Mitigation}.
\newblock {\em Expert Systems with Applications}, 38(10):13351--13359, 2011.

\bibitem{kairouz2019}
P.~Kairouz, H.~B. McMahan, B.~Avent, A.~Bellet, M.~Bennis, A.~N. Bhagoji,
  K.~Bonawitz, Z.~Charles, G.~Cormode, R.~Cummings, et~al.
\newblock {Advances and Open Problems in Federated Learning}.
\newblock {\em arXiv preprint arXiv:1912.04977}, 2019.

\bibitem{kang2020}
Y.~Kang, Y.~Liu, L.~Ding, X.~Liu, X.~Tong, and W.~Wang.
\newblock Differentially private erm based on data perturbation.
\newblock {\em arXiv preprint arXiv:2002.08578}, 2020.

\bibitem{kawa2019}
D.~Kawa, S.~Punyani, P.~Nayak, A.~Karkera, and V.~Jyotinagar.
\newblock Credit risk assessment from combined bank records using federated
  learning.
\newblock {\em International Research Journal of Engineering and Technology
  (IRJET)}, 6(4):1355--1358, 2019.

\bibitem{Khan2009}
R.~Khan, M.~Corney, A.~Clark, and G.~Mohay.
\newblock A role mining inspired approach to representing user behaviour in erp
  systems.
\newblock In {\em Asia Pacific Industrial Engineering \& Management Systems
  Conference 2009}. APIEMS Society, 2009.

\bibitem{kifer2012}
D.~Kifer, A.~Smith, and A.~Thakurta.
\newblock Private convex empirical risk minimization and high-dimensional
  regression.
\newblock In {\em Conference on Learning Theory}, pages 25--1. JMLR Workshop
  and Conference Proceedings, 2012.

\bibitem{kim2018}
H.~Kim, J.~Park, M.~Bennis, and S.-L. Kim.
\newblock On-device federated learning via blockchain and its latency analysis.
\newblock {\em arXiv preprint arXiv:1808.03949}, 2018.

\bibitem{kogan2021}
A.~Kogan and C.~Yin.
\newblock Privacy-preserving information sharing within an audit firm.
\newblock {\em Journal of Information Systems}, 35(2):243--268, 2021.

\bibitem{konevcny2016}
J.~Kone{\v{c}}n{\`y}, H.~B. McMahan, D.~Ramage, and P.~Richt{\'a}rik.
\newblock Federated optimization: Distributed machine learning for on-device
  intelligence.
\newblock {\em arXiv preprint arXiv:1610.02527}, 2016.

\bibitem{lecun2015}
Y.~LeCun, Y.~Bengio, and G.~Hinton.
\newblock Deep learning.
\newblock {\em {Nature}}, 521(7553), 2015.

\bibitem{mathews2022}
S.~M. Mathews and S.~Assefa.
\newblock {Federated Learning: Balancing the Thin Line Between Data
  Intelligence and Privacy}.
\newblock {\em AAAI Workshop on AI in Financial Services: Adaptiveness,
  Resilience \& Governance}, 2022.

\bibitem{McGlohon2009}
M.~McGlohon, S.~Bay, M.~G.~M. Anderle, D.~M. Steier, and C.~Faloutsos.
\newblock {SNARE: A Link Analytic System for Graph Labeling and Risk
  Detection}.
\newblock In {\em KDD'09: 15th ACM SIGKDD Conference on Knowledge Discovery and
  Data Mining}, 2009.

\bibitem{mcmahan2017a}
B.~McMahan, E.~Moore, D.~Ramage, S.~Hampson, and B.~A. y~Arcas.
\newblock {Communication-Efficient Learning of Deep Networks from Decentralized
  Data}.
\newblock In {\em Artificial Intelligence and Statistics}, pages 1273--1282.
  PMLR, 2017.

\bibitem{munoko2020}
I.~Munoko, H.~L. Brown-Liburd, and M.~Vasarhelyi.
\newblock {The Ethical Implications of Using Artificial Intelligence in
  Auditing}.
\newblock {\em Journal of Business Ethics}, 167(2), 2020.

\bibitem{nonnenmacher2021b}
J.~Nonnenmacher and J.~M. G{\'o}mez.
\newblock Unsupervised anomaly detection for internal auditing: Literature
  review and research agenda.
\newblock {\em The International Journal of Digital Accounting Research},
  21(27):1--22, 2021.

\bibitem{nonnenmacher2021a}
J.~Nonnenmacher, F.~Kruse, G.~Schumann, and J.~Marx~G{\'o}mez.
\newblock Using autoencoders for data-driven analysis in internal auditing.
\newblock In {\em Proceedings of the 54th Hawaii International Conference on
  System Sciences}, page 5748, 2021.

\bibitem{papernot2016}
N.~Papernot, M.~Abadi, U.~Erlingsson, I.~Goodfellow, and K.~Talwar.
\newblock Semi-supervised knowledge transfer for deep learning from private
  training data.
\newblock {\em arXiv preprint arXiv:1610.05755}, 2016.

\bibitem{ramamurthy2021}
R.~Ramamurthy, M.~Pielka, R.~Stenzel, C.~Bauckhage, R.~Sifa, T.~D. Khameneh,
  U.~Warning, B.~Kliem, and R.~Loitz.
\newblock Alibert: improved automated list inspection (ali) with bert.
\newblock In {\em 21st ACM Symposium on Document Engineering}, 2021.

\bibitem{rivest1978}
R.~L. Rivest, L.~Adleman, M.~L. Dertouzos, et~al.
\newblock {On Data Banks and Privacy Homomorphisms}.
\newblock {\em Foundations of secure computation}, 4(11):169--180, 1978.

\bibitem{schreyer2022}
M.~Schreyer, M.~Baumgartner, F.~Ruud, and D.~Borth.
\newblock {Artificial Intelligence in Internal Audit as a Contribution to
  Effective Governance}.
\newblock {\em Expert Focus}, (01), 2022.

\bibitem{schreyer2021}
M.~Schreyer, T.~Sattarov, and D.~Borth.
\newblock Multi-view contrastive self-supervised learning of accounting data
  representations for downstream audit tasks.
\newblock In {\em International Conference on Artificial Intelligence}, 2021.

\bibitem{schreyer2017}
M.~Schreyer, T.~Sattarov, D.~Borth, A.~Dengel, and B.~Reimer.
\newblock Detection of anomalies in large scale accounting data using deep
  autoencoder networks.
\newblock {\em arXiv preprint arXiv:1709.05254}, 2017.

\bibitem{schreyer2020}
M.~Schreyer, T.~Sattarov, A.~S. Gierbl, B.~Reimer, and D.~Borth.
\newblock Learning sampling in financial statement audits using vector
  quantised variational autoencoder neural networks.
\newblock In {\em International Conference on Artificial Intelligence}, 2020.

\bibitem{schreyer2019b}
M.~Schreyer, T.~Sattarov, B.~Reimer, and D.~Borth.
\newblock Adversarial learning of deepfakes in accounting.
\newblock {\em NeurIPS 2019 Workshop on Robust AI in Financial Services,
  Vancouver, BC, Canada}, 2019.

\bibitem{schreyer2019a}
M.~Schreyer, T.~Sattarov, C.~Schulze, B.~Reimer, and D.~Borth.
\newblock Detection of accounting anomalies in the latent space using
  adversarial autoencoder neural networks.
\newblock {\em 2nd KDD Workshop on Anomaly Detection in Finance, USA}, 2019.

\bibitem{schultz2020}
M.~Schultz and M.~Tropmann-Frick.
\newblock Autoencoder neural networks versus external auditors: Detecting
  unusual journal entries in financial statement audits.
\newblock In {\em Proceedings of the 53rd Hawaii International Conference on
  System Sciences}, 2020.

\bibitem{sifa2019}
R.~Sifa, A.~Ladi, M.~Pielka, R.~Ramamurthy, L.~Hillebrand, B.~Kirsch,
  D.~Biesner, R.~Stenzel, T.~Bell, M.~L{\"u}bbering, et~al.
\newblock Towards automated auditing with machine learning.
\newblock In {\em ACM Symposium on Document Engineering 2019}, 2019.

\bibitem{smith2017}
V.~Smith, C.-K. Chiang, M.~Sanjabi, and A.~S. Talwalkar.
\newblock Federated multi-task learning.
\newblock {\em Advances in neural information processing systems}, 30, 2017.

\bibitem{sun2019}
T.~Sun.
\newblock {Applying Deep Learning to Audit Procedures: An Illustrative
  Framework}.
\newblock {\em Accounting Horizons}, 33(3):89--109, 2019.

\bibitem{sun2017}
T.~Sun and M.~A. Vasarhelyi.
\newblock Deep learning and the future of auditing: How an evolving technology
  could transform analysis and improve judgment.
\newblock {\em CPA Journal}, 87(6), 2017.

\bibitem{thiprungsri2011}
S.~Thiprungsri and M.~A. Vasarhelyi.
\newblock {Cluster Analysis for Anomaly Detection in Accounting Data: An Audit
  Approach}.
\newblock {\em International Journal of Digital Accounting Research}, 11, 2011.

\bibitem{vincent2008}
P.~Vincent, H.~Larochelle, Y.~Bengio, and P.-A. Manzagol.
\newblock Extracting and composing robust features with denoising autoencoders.
\newblock In {\em Proceedings of the 25th international conference on Machine
  learning}, pages 1096--1103, 2008.

\bibitem{yao1982}
A.~C. Yao.
\newblock {Protocols for Secure Computations}.
\newblock In {\em {23rd Annual Symposium on Foundations of Computer Science}},
  pages 160--164. IEEE, 1982.

\bibitem{yin2021}
H.~Yin, A.~Mallya, A.~Vahdat, J.~M. Alvarez, J.~Kautz, and P.~Molchanov.
\newblock See through gradients: Image batch recovery via gradinversion.
\newblock In {\em Proceedings of the IEEE/CVF Conference on Computer Vision and
  Pattern Recognition}, 2021.

\bibitem{yoon2015}
K.~Yoon, L.~Hoogduin, and L.~Zhang.
\newblock {Big Data as Complementary Audit Evidence}.
\newblock {\em {Accounting Horizons}}, 29(2):431--438, 2015.

\bibitem{yousefpour2021}
A.~Yousefpour, I.~Shilov, A.~Sablayrolles, D.~Testuggine, K.~Prasad, M.~Malek,
  J.~Nguyen, S.~Ghosh, A.~Bharadwaj, J.~Zhao, G.~Cormode, and I.~Mironov.
\newblock Opacus: {U}ser-friendly differential privacy library in {PyTorch}.
\newblock {\em arXiv preprint arXiv:2109.12298}, 2021.

\bibitem{yunis2021}
M.~M. Yunis, R.~El-Khalil, and M.~Ghanem.
\newblock {Towards a Conceptual Framework on the Importance of Privacy and
  Security Concerns in Audit Data Analytics}.
\newblock 2021.

\bibitem{yurochkin2019}
M.~Yurochkin, M.~Agarwal, S.~Ghosh, K.~Greenewald, N.~Hoang, and Y.~Khazaeni.
\newblock Bayesian nonparametric federated learning of neural networks.
\newblock In {\em International Conference on Machine Learning}, pages
  7252--7261. PMLR, 2019.

\bibitem{zhang2017}
J.~Zhang, K.~Zheng, W.~Mou, and L.~Wang.
\newblock Efficient private erm for smooth objectives.
\newblock {\em arXiv preprint arXiv:1703.09947}, 2017.

\bibitem{zheng2020}
Y.~Zheng, Z.~Wu, Y.~Yuan, T.~Chen, and Z.~Wang.
\newblock Pcal: A privacy-preserving intelligent credit risk modeling framework
  based on adversarial learning.
\newblock {\em arXiv preprint arXiv:2010.02529}, 2020.

\bibitem{zupan2020}
M.~Zupan, V.~Budimir, and S.~Letinic.
\newblock Journal entry anomaly detection model.
\newblock {\em Intelligent Systems in Accounting, Finance and Management},
  27(4):197--209, 2020.

\end{thebibliography}

\end{document}